\RequirePackage{fix-cm}
\documentclass[twocolumn]{svjour3}          
\smartqed  
\usepackage{booktabs} 
\usepackage{graphicx}
\usepackage{graphics}
\usepackage{amsmath}
\usepackage{amsfonts}
\usepackage{algorithm}
\usepackage{algorithmic}
\usepackage{url}
\usepackage{float}
\usepackage{multirow} 
\usepackage{comment}
\usepackage{subcaption}
\usepackage{mathrsfs}  
\usepackage{cite}
\usepackage{xcolor}
\newcommand{\JCN}{\mathrm{JCN}}
\usepackage[labelfont=bf]{caption}
\allowdisplaybreaks

\newcommand{\figref}[1]{Figure \ref{#1}}

\newcommand{\secref}[1]{Section~\ref{#1}}

\newcommand{\etal}{et~al. }

\newcommand{\M}{\mathbb{M}}
\newcommand{\cM}{\mathcal{M}}
\newcommand{\cS}{\mathbb{S}}
\newcommand{\R}{\mathbb{R}}

\newcommand{\Z}{\mathbb{Z}}
\newcommand{\RG}{\mathcal{RG}}
\newcommand{\RS}{\mathcal{RS}}
\newcommand{\ExPD}{ExPD}
\newcommand{\MDRG}{\mathbb{MR}}

\newcommand{\cO}{\mathcal{O}}

\newcommand{\desc}[1]{\left|\hat{\phi}_{#1}\right|}
\newcommand{\bc}{\mathbf{c}}
\newcommand{\f}{\mathbf{f}}
\newcommand{\g}{\mathbf{g}}
\newcommand{\h}{\mathbf{h}}
\newcommand{\x}{\mathbf{x}}

\newcommand{\PD}{\mathrm{PD}}
\newcommand{\Dg}{\mathrm{Dg}}
\newcommand{\ExDg}{\mathrm{ExDg}}
\newcommand{\Range}{R}
\newcommand{\abs}[1]{\lvert{#1}\rvert}

\begin{document}

\title{A Topological Distance Measure between  Multi-Fields for Classification and Analysis of Shapes and  Data 
}


\author{Yashwanth Ramamurthi       \and
        Amit Chattopadhyay 
}


 \institute{Yashwanth Ramamurthi \at
             International Institute of Information Technology, Bangalore \\
             \email{yashwanth@iiitb.ac.in}           
            \and
            Amit Chattopadhyay \at
             International Institute of Information Technology, Bangalore\\
             \email{a.chattopadhyay@iiitb.ac.in}
 }
\date{}


\maketitle

\begin{abstract}
Distance measures play an important role in shape classification and data analysis problems. Topological distances based on Reeb graphs and persistence diagrams have been employed to obtain effective algorithms in shape matching and scalar data analysis. In the current paper, we propose an improved distance measure between two multi-fields  by computing a multi-dimensional Reeb graph (MDRG) each of which captures the topology of a multi-field through a hierarchy of Reeb graphs in different dimensions. A hierarchy of persistence diagrams is then constructed by computing a persistence diagram corresponding to  each Reeb graph of the MDRG. Based on this representation, we propose  a novel distance measure between two MDRGs by extending the bottleneck distance between two Reeb graphs. We show that the proposed measure satisfies the pseudo-metric and stability properties. We examine the effectiveness of the proposed multi-field topology-based measure on two different applications: (1) shape classification and (2) detection of topological features in a time-varying multi-field data. In the shape classification problem, the performance of the proposed measure is compared with the well-known topology-based measures in shape matching. In the second application, we consider a time-varying volumetric multi-field data from the field of computational chemistry where the goal is to detect the site of stable bond formation between Pt and CO molecules. We demonstrate the ability of the proposed distance in classifying each of the sites as occurring before and after the bond stabilization.

\end{abstract}
\sloppy
\section{Introduction}
The computations of topological similarity between shapes or data reveal critical phenomena to experts in various domains.  A lot of research has been done in developing similarity or distance measures using scalar topology. The development of these measures were based on descriptors in scalar topology such as Reeb graph, Morse-Smale Complex, extremum graph and persistence diagram. Scalar topology based methods have found applications in several areas such as  shape matching \cite{2001-Hilaga-MRG, 2007-Tam-DeformableModelRetrieval}, protein structure classification \cite{2004-Zhang-Dual-Contour-Tree}, symmetry detection \cite{2014-Thomas-Multiscale} and periodicity detection in time-varying data \cite{2018-Sridharamurthy-Edit-Distance-Merge-Trees}.

Persistent homology theory \cite{2002-Edelsbrunner-Persistence, 2007-Zomorodian-ComputingPersistentHomology, 2009-Chazal-ProximityOfPersistenceModules} has provided a simple yet powerful technique to encode the topological information of a scalar field into a bar code or persistence diagram. Distance measures based on persistence diagrams have been shown to be effective in the classification of signals \cite{2017-Marchese-Signal-Classification} and determining the crystal structure of materials \cite{2020-Maroulas-Distance-Topological-Classification}. The bottleneck distance between persistence diagrams has proven to be effective in the topological analysis of shapes and data \cite{2014-Li-Persistence-Based-Structural-Recognition, 2016-MultiscaleMapper}. Recently, Bauer \etal \cite{2014-Bauer-DistanceBetweenReebGraphs} developed a functional distortion distance between Reeb graphs and showed that the bottleneck distance between the persistence diagrams of Reeb graphs is a pseudo-metric.

Multi-field topology is getting wider attention because of its ability to capture and classify richer topological features than scalar topology, as has been shown in various data analysis applications in computational physics and computational chemistry \cite{2012-Duke-Nuclear-Scission, 2019-Agarwal-histogram, 2021-Ramamurthi-MRS}. However, the application of multi-field topology in the classification and analysis of shapes and data, requires further development in the  representation of such topological features and finding effective distance measures between such representations. In the current work, we propose a novel distance measure between two shapes or multi-fields based on their multi-dimensional Reeb graphs (MDRG).

The MDRG is a topological descriptor which captures the topology of a multi-field using a series of Reeb graphs in different dimensions. To compare two MDRGs, we compute the persistence diagrams corresponding to the component Reeb graphs and find a distance between two such collections of Reeb graphs using the bottleneck distance. We show that the proposed distance measure is stable and satisfies the pseudo-metric property. We validate the effectiveness of the proposed distance measure in the classification of shapes and detection of topological features in time-varying volumetric data. In the current paper, our contributions are as follows.
\begin{itemize}
    \item We propose a novel distance measure between two MDRGs based on the bottleneck distance between the component Reeb graphs of the MDRGs. 
    
    \item We show the proposed distance measure satisfies the stability and pseudo-metric properties.
    
    \item We show the effectiveness of using pairs of eigenfunctions from the Laplace-Beltrami operator, in the classification of shapes, compared to other well-known shape descriptors, viz. HKS, WKS and SIHKS, for different topology-based techniques, using the SHREC $2010$ dataset.
    
    \item We demonstrate the performance of multi-field topology over scalar topology by showing the effectiveness of the proposed measure in detecting stable bond formation in the Pt-CO molecular orbitals data and classifying sites based on their occurrence before and after bond stabilization.
\end{itemize}

\noindent
\textbf{Outline.} In \secref{sec:related-work}, we discuss topological tools and their applications in shape matching and data analysis. \secref{sec:background} discusses the necessary background required to understand the proposed distance measure. In \secref{section:distance-measure-between-mdrgs}, we describe the proposed distance measure between MDRGs and prove its pseudometric and stability properties. In \secref{sec:experimental-results}, we show the experimental results of the proposed measure in (i) shape classification and (ii) topological feature detection in a time-varying dataset from computational chemistry. Finally, in \secref{sec:conclusion} we draw a conclusion and provide future directions.

\section{Related Work}
\label{sec:related-work}
Several shape classification and data analysis techniques have been developed based on the scalar topology tools such as contour Tree \cite{2004-Zhang-Dual-Contour-Tree}, Reeb Graph \cite{2001-Hilaga-MRG}, and Merge tree \cite{2018-Sridharamurthy-Edit-Distance-Merge-Trees}. Li \etal \cite{2014-Li-Persistence-Based-Structural-Recognition} computed the persistence diagrams of shapes corresponding to the spectral descriptors HKS, WKS, SIHKS and compared the technique with the bag-of-features model. Kleiman \etal \cite{2017-Kleiman-StructureBasedShapeCorrespondence} present a method for computing region-level correspondence between a pair of shapes by obtaining a shape graph corresponding to each shape and then matching the nodes of shape graphs. Poulenard \etal \cite{2018-Poulenard-TopologicalFunctionOptimization} present a characterization of functional maps based on persistence diagrams along with an optimization scheme to improve the computational efficiency.

An increase of interest in the Laplace-Beltrami (LB) operator resulted in the evolution of various signatures to capture the features in a shape \cite{1997-Rosenberg-Laplacian, 2009-Reuter-LaplaceBeltrami}. Reuter \etal \cite{2006-Reuter-Shape-DNA} proposed the shape DNA, where a shape is represented using the eigenvalues of the Laplace-Beltrami operator and was able to identify the shapes with similar poses. However, non-isometric shapes can have the same set of eigenvalues. This drawback is overcome by the Global Point Signature (GPS) \cite{2007-Rustamov-GPS}, where the eigenfunctions are used along with the eigenvalues. However, this introduced the sign ambiguity problem to the eigenfunctions (see \secref{subsubsec:feature-descriptors-of-shape} for more details), which was solved using the Heat Kernel Signature (HKS) \cite{2009-Sun-HKS}. However, the HKS consists of information from low frequencies. The suppression of high frequencies makes it difficult to detect microscopic features. This limitation was overcome in the Wave Kernel Signature (WKS)\cite{2011-Aubry-WKS} by using band-pass filters instead of low-pass filters. Bronstein \etal \cite{2010-Bronstein-SIHKS} propose a scale invariant version of the Heat Kernel Signature (SIHKS) by using logarithmic sampling and Fourier transform. Recently, Zihao \etal \cite{2020-Wang-Shape-Retrieval} proposed a shape descriptor based on the probability distributions of the  eigenfunctions of the LB operator and showed its effectiveness over SIHKS. In this paper, we measure distance between shapes by directly comparing the eigenfunctions of the LB operator and show its performance with respect to HKS, WKS and SIHKS.

Recently, tools for capturing multi-field topology have been studied using the Jacobi set \cite{2004-Edelsbrunner-JacobiSets}, Reeb Space  \cite{2008-Edelsbrunner-ReebSpaces}, Mapper \cite{2007-Singh-Mapper}, Joint Contour Net \cite{2014-Carr-JCN}, Multi-Dimensional Reeb Graph \cite{2014-Chattopadhyay-ExtractingJacobiStructures, 2016-Chattopadhyay-MultivariateTopologySimplification}, etc.  Techniques for comparing multi-fields have been developed such as the bottleneck distance between persistence diagrams of multiscale mappers \cite{2016-MultiscaleMapper}, distance between fiber-component distributions \cite{2019-Agarwal-histogram} and similarity between multi-resolution Reeb spaces (MRSs) \cite{2021-Ramamurthi-MRS}. In the current paper, we capture the topological features in a multi-field by computing the persistence diagrams of the Reeb graphs in an MDRG and develop a distance measure between two such MDRGs based on the bottleneck distance between the component Reeb graphs of the MDRGs. We evaluate the effectiveness of the proposed distance measure in (i) the shape classification problem using the SHREC $2010$ dataset \cite{2010-SHREC} and (ii) analysis and classification of topological features in a volumetric time-varying data from the field of computational chemistry.
\section{Background}
\label{sec:background}

In this section, we discuss the necessary background required to understand the proposed distance measure and feature descriptors used in the comparison of shapes.

\subsection{Multi-Field Topology}
\label{sec:multi-jcn}
Let $\mathbb{M}$ be a triangulated mesh of a compact $d$-manifold $\mathcal{M}$. A multi-field on $\M$ with $r$ component scalar fields is defined as a continuous map $\f = (f_1, f_2,...,f_r) : \mathbb{M} \rightarrow \mathbb{R}^r$. For a value $\bc \in \mathbb{R}^r$, its inverse $\f^{-1}(\bc)$ is called a \emph{fiber}. Each connected component of a fiber is a \emph{fiber-component} \cite{2004-Saeki-Topology-of-Singular-Fibers, 2014-Saeki-Visualizing-Multivariate-Data}. In particular, for a scalar field $f : \mathbb{M} \rightarrow \mathbb{R}$, the inverse of $f$ corresponding to a point in the range is called a \emph{level set} and each connected component of a level set is called a \emph{contour}. The Reeb space of $\f$, denoted by $\RS_{\f}$, is the quotient space by contracting each fiber-component to a point \cite{2008-Edelsbrunner-ReebSpaces}. In particular, the Reeb space of a scalar field $f : \mathbb{M} \rightarrow \mathbb{R}$ is known as the Reeb graph $\RG_f$, which is obtained by contracting each contour to a point and $\tilde{f}: \mathbb{M} \rightarrow \RG_{f}$ is the corresponding quotient map.  \cite{2017-Saeki-Theory-of-Singular-Fibers}. Carr \etal \cite{2014-Carr-JCN} proposed the Joint Contour Net (JCN) data-structure, which is a quantized approximation of the Reeb space. For construction of the JCN of $\f$ , the range of $\f$ is subdivided into a finite set of $r$-dimensional intervals. For each interval, the inverse of $\f$ is a \emph{quantized fiber} and each connected component of a quantized fiber is a \emph{quantized fiber-component} or \emph{joint contour}. The JCN is a graph where a node corresponds to a joint contour and an edge between two nodes corresponds to the adjacency of the corresponding joint contours in the domain.

\subsection{Multi Dimensional Reeb Graph}
\label{background:mdrg}
\begin{figure*}[!ht]
        \centering
        \includegraphics[width=0.8\textwidth]{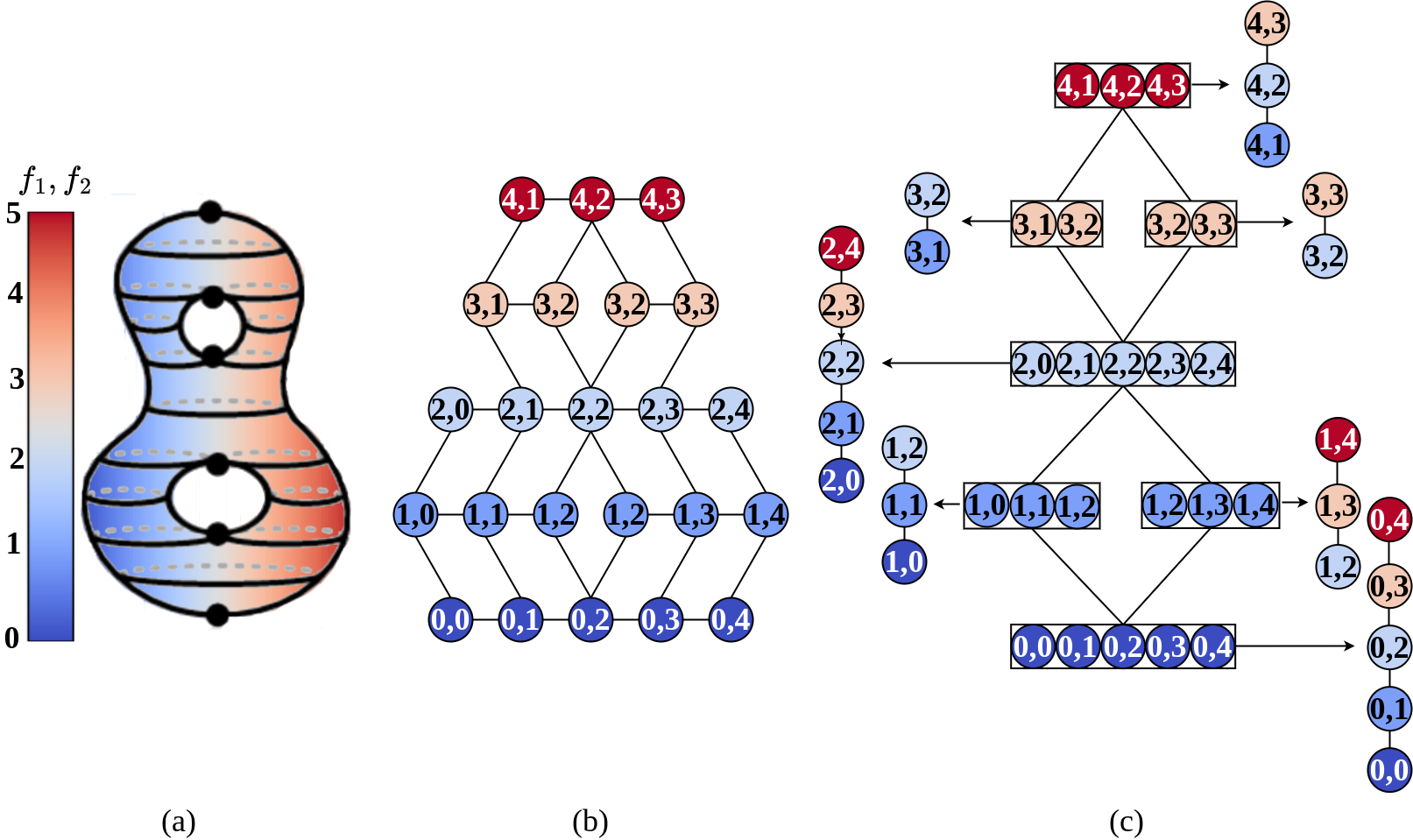}
        \caption{(a) A piecewise linear bivariate data over a double torus: The black curves depict the contours of the first field $(f_1)$ and the mesh is colored based on the values of the second field $(f_2)$, (b) JCN at $5\times 5$ resolution: the coloring of the nodes are is based on the values of the first field, (c) MDRG computed using the algorithm in \cite{2016-Chattopadhyay-MultivariateTopologySimplification}: the coloring of the nodes in the first (second) dimension is based on the values of $f_1$ ($f_2$)}
        \label{fig:JCN_MDRG}
\end{figure*}

A multi-dimensional Reeb graph (MDRG) of a multi-field $\f = (f_1, ..., f_r) : \mathbb{M} \rightarrow \mathbb{R}^r$, proposed by Chattopadhyay \etal \cite{2014-Chattopadhyay-ExtractingJacobiStructures, 2016-Chattopadhyay-MultivariateTopologySimplification}, is the decomposition of the Reeb space (or a JCN) into a series of Reeb graphs in different dimensions. To construct the MDRG of $\f$, the Reeb graph $\RG_{f_1}$ of the function $f_1$ is computed.  Each point $p \in \RG_{f_1}$ represents a contour $C_p$ of $f_1$. For every $p \in \RG_{f_1}$ a Reeb graph corresponding to the field $f_2$ is computed by restricting $f_2$ on $C_p$. This procedure is repeated by computing the Reeb graphs for the function $f_i$ by restricting $f_i$ on the contours corresponding to points in the Reeb graphs of $f_{i-1}$, where $2 \leq i \leq r$. In practice, the MDRG is constructed from a quantized approximation of the Reeb Space (JCN). Therefore, similar to the JCN, the MDRG also depends on the subdivision of the range. Each node in a Reeb graph of the MDRG corresponds to a \emph{quantized contour} and the adjacency between quantized contours is represented by an edge in the Reeb graph. \figref{fig:JCN_MDRG} shows the JCN and MDRG corresponding to a bivariate field. In the current paper, we compute persistence diagrams corresponding to Reeb graphs in the MDRG and propose a distance to compare the collection of persistence diagrams of two MDRGs based on the bottleneck distance between Reeb graphs \cite{2014-Bauer-DistanceBetweenReebGraphs}. 

\subsection{Persistence Diagram}
In this paper, we give a brief introduction to the notion of \emph{persistence diagrams} and refer the readers to \cite{2010-Edelsbrunner-book, 2007-Zomorodian-ComputingPersistentHomology} for further details on persistent homology.

Let $f$ be a continuous real-valued function defined on $\M$. For a value $a \in \mathbb{R}$, the sublevel set $\M_{\leq a}$ consists of the points in $\mathbb{M}$ with $f$-value less than or equal to $a$, i.e. $\M_{\leq a} = f^{-1}(-\infty, a]$.  For $a \leq b$, the $l$-th homology groups of the sublevel sets $\M_{\leq a}$ and $\M_{\leq b}$ are connected by the inclusion map $f^{a,b}_l : H_l(\M_{\leq a}) \rightarrow H_l(\M_{\leq b})$. A value $a \in \mathbb{R}$ is a \emph{homological critical value} of $f$ if $\exists l \in \Z^{*}$ such that $f^{a-\delta, a + \delta}_l$ is not an isomorphism for all sufficiently small $\delta > 0$, where $\Z^{*}$ is the set of non-negative integers. We assume that $f$ is \emph{tame}, i.e, the number of homological critical values of $f$ is finite and the homology groups $H_l(\M_{\leq a})$ are finite-dimensional $\forall l \in \Z^{*}$. Here, $a$ is any homological critical value of $f$. Let $a_0 < a_1 < ... < a_N$ be the homological critical values of $f$.  In this paper, we consider homology with coefficients in $\mathbb{Z}_2$, which is the group of integers modulo $2$. Therefore,  $H_l(\M_{\leq _{a_i}})$ is a vector space for $0 \leq i \leq N$. We have the following sequence of vector spaces,
\begin{small}
\begin{equation}
\emptyset = H_l(\M_{\leq a_0}) \rightarrow H_l(\M_{\leq a_1}) \rightarrow \cdots \rightarrow H_l(\M_{\leq a_N}) =  H_l(\M).
    \label{eqn:homology-sequence-ordinary}
\end{equation}
\end{small}
where the homomorphisms $f_l^{a_i,a_{i+1}}$ are induced by the inclusions $\M_{\leq a_i} \subseteq \M_{\leq a_{i+1}}$. A homology class $\gamma$ is born at $a$ if $\gamma \in H_l(\M_{a})$ but $\gamma \notin \text{Im } f_l^{a-\delta,a}$ for $0 < \delta \leq b-a$. Further, a class $\gamma$ which is born at $a$ dies at $b$ if $f_l^{a,b-\delta}(\gamma) \notin \text{Im } f_l^{a-\delta,b-\delta}$ for any $\delta > 0$, but $f_l^{a,b}(\gamma) \in \text{Im } f_l^{a-\delta,b}$. Such birth and death events are recorded by persistent homology. The $l$-th ordinary persistence diagram is a multiset of points in $\mathbb{\overline{R}}^2$, where $\mathbb{\overline{R}} = \mathbb{R} \cup \{-\infty, + \infty \}$. Each point $(a,b)$ in the $l$th ordinary persistence diagram corresponds to a $l$-homology class which is born at $a$ and dies at $b$. The multiplicity of a point $(a,b)$ with $a \leq b$ is defined in terms of the ranks of the homomorphism $f_{l}^{a,b}$ and the points along the diagonal have infinite multiplicity.

In general, not all homology classes die during the sequence in equation (\ref{eqn:homology-sequence-ordinary}). Such homology classes are is said to be essential. An essential homology class of dimension $l$ is denoted by a point $(a_i, \infty)$ in the $l$th ordinary persistence diagram, where $a_i$ is the critical value corresponding to the birth of the homology class. By appending a sequence of relative homology groups to equation (\ref{eqn:homology-sequence-ordinary}), we obtain the following sequence:
\begin{small}
\begin{align}
    \nonumber \emptyset &= H_l(\M_{\leq a_0}) \rightarrow H_l(\M_{\leq a_1}) \rightarrow \cdots \rightarrow H_l(\M_{\leq a_N}) 
 =  H_l(\M)\\
  &=H_l(\M,\M_{\geq a_N}) \rightarrow H_l(\M,\M_{\geq a_{N-1}}) \rightarrow \cdots \nonumber\\
&\hspace{0.5cm}\cdots\rightarrow H_l(\M,\M_{\geq a_0}) = \emptyset. \label{eqn:homology-sequence-extended} 
\end{align}
\end{small}
where $\M_{\geq a_i}$ denotes the super-level set of $f$, $\M_{\geq a_i} = f^{-1}[a_i,\infty)$ and $H_l(\M,\M_{\geq a_i})$ is a relative homology group \cite{2010-Edelsbrunner-book}. Essential homology classes are created in the ordinary part and are destroyed in the relative part of the sequence in equation (\ref{eqn:homology-sequence-extended}). The birth and death of essential homology classes are encoded by the extended persistence diagram. An essential homology class of dimension $l$ which is born at $H_l(\M_{\leq a_i})$ and dies at $ H_l(\M,\M_{\geq a_j})$ is denoted by the point $(a_i, a_j)$ in the $l$th extended persistence diagram.

\subsection{Bottleneck Distance between Persistence Diagrams}
\label{subsec:bottleneck-distance-between-persistence-diagrams}
The bottleneck distance between persistence diagrams $X$ and $Y$ is defined as follows:
\begin{small}
\begin{equation}
    d_B(X,Y) = \inf_{\eta : X \rightarrow Y} \max_{x \in X} \|\x - \eta(x)\|_{\infty}
\end{equation}
\end{small}
where $\eta$ ranges over bijections between $X$ and $Y$. For each point $(a,b)$ in $X$, we add its nearest diagonal point $(\frac{a+b}{2}, \frac{a+b}{2})$ in $Y$ and vice-versa. The number of points in $X$ and $Y$ are now equal. To compute the bottleneck distance, a bijection $\eta : X \rightarrow Y$ is constructed such that $\underset{x \in X}{\max} \|x - \eta(x)\|_{\infty}$ is minimized.

\subsection{Reeb Graph and Persistence Diagram}
\label{sec:persistence-diagram-of-reeb-graph}

Let $f$ be a real-valued continuous function defined on $\M$. The Reeb graph of $f$, denoted by $\RG_{f}$, is the quotient space of contours of $f$. The function $f$ induces a real-valued function $\bar{f} : \RG_{f} \rightarrow \R$, which takes each point in $p \in \RG_{f}$ to the value of $f$ corresponding to its contour $\tilde{f}^{-1}(p)$ in the domain, $f = \bar{f} \circ \tilde{f}$.

The persistence of topological features of $\RG_{f}$ are encoded in the persistence diagrams  $\Dg_0(\RG_{f})$, $\Dg_0(\RG_{-f})$, $\ExDg_0(\RG_{f})$ and  $\ExDg_1(\RG_{f})$. The $0$-dimensional persistent homological features corresponding to the sub-level set and super-level set filtrations of $f$ are captured in $\Dg_0(\RG_{f})$ and $\Dg_0(\RG_{-f})$ respectively. $\ExDg_0(\RG_{f})$ encodes the range of $f$ and $\ExDg_1(\RG_{f})$ captures the $1$-cycles or loops in $\RG_{f}$.

To compute the persistence diagrams of $\RG_{f}$, we require $\RG_f$ to be the Reeb graph of a Morse function. If $f$ is a \emph{Morse function}, i.e. all its critical points are non-degenerate and are at different levels, then the critical nodes of $\RG_f$ have distinct values of $\bar{f}$ and belong to one of the following five-types: (i) a minimum (with down-degree $= 0$, up-degree $= 1$), (ii) a maximum (with up-degree $= 0$, down-degree $ = 1$), (iii) a down-fork (with down-degree $= 2$, up-degree $ = 1$) and (iv) a up-fork (with up-degree $= 2$, down-degree $= 1$).  A regular node has up-degree = $1$ and down-degree $= 1$. A down-fork (similarly, up-fork) node is called an essential down-fork node when it contributes to a loop (cycle) of the Reeb graph. Otherwise it is called an ordinary down-fork node. To ensure that $\RG_f$ is the Reeb graph of a Morse function, we first eliminate degenerate critical nodes in $\RG_{f}$ by breaking them into non-degenerate critical nodes \cite{2019-Tu-PropogateAndPair}. After eliminating degenerate critical nodes, we ensure that the critical nodes of $\bar{f}$ are at different levels. If two critical nodes are at the same level, then the value of one of the nodes is increased/decreased by a small value $\epsilon$. After removing degenerate critical nodes and ensuring that critical nodes are at different levels, $\RG_{f}$ becomes the Reeb graph of a Morse function.

\begin{figure}
    \centering
    \includegraphics[width=0.45\textwidth]{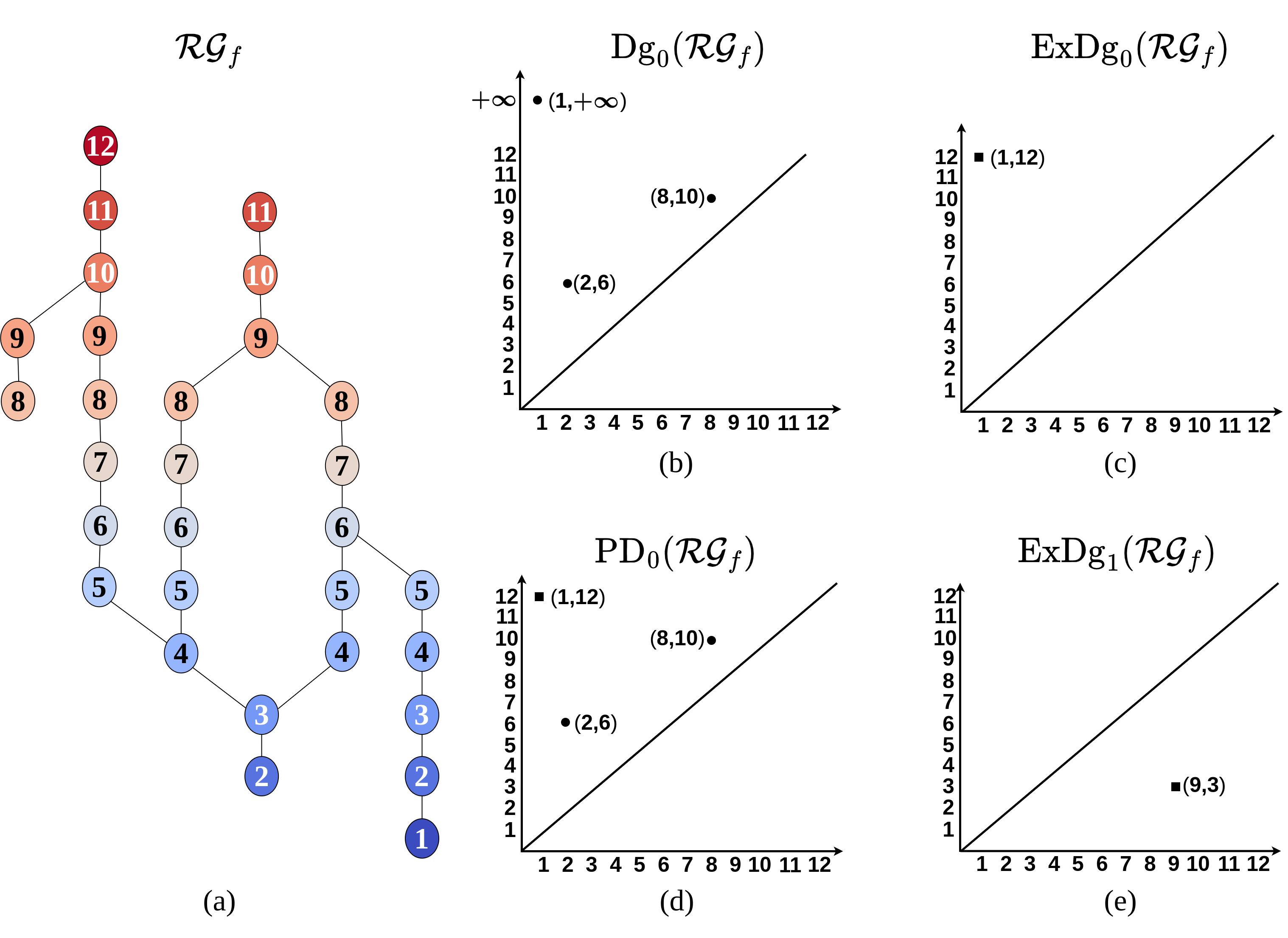}
    \caption{(a) Reeb graph of a scalar field $f$. (b) $0$th ordinary persistence diagram $\Dg_0(\RG_{f})$. (c) $0$th extended persistence diagram $\ExDg_0(\RG_{f})$. (d) $\PD_0(\RG_f):=\Dg_0(\RG_f)\cup \ExDg_0(\RG_f)\setminus \{(1,\infty)\}$. (e) $1$st extended persistence diagram $\ExDg_1(\RG_{f})$. The points in ordinary (extended) persistence diagrams are denoted by circular points (squares)}
    \label{fig:persistence-diagram-Reeb-graph}
\end{figure}

The points in $\Dg_0(f)$ are computed by pairing ordinary down-forks with minima, ordinary up-forks with maxima and the global minimum with global maximum. Let $u$ be an ordinary down-fork of $\RG_{f}$. The two lower branches of $u$ correspond to two different components $C_1$ and $C_2$ in $(\RG_{f})_{\leq \bar{f}(u)}$. Let $x_1$ and $x_2$ be the global minimum of $C_1$ and $C_2$ respectively. Let us assume that $\bar{f}(x_1) < \bar{f}(x_2)$. Then a $0$-dimensional homology class is born at $x_2$ and dies at $u$. $x_2$ is paired with $u$ and point $(\bar{f}(x_2), \bar{f}(u))$ is created in $\Dg_0(\RG_{f})$ (see \figref{fig:persistence-diagram-Reeb-graph}(b)). A symmetric procedure is applied on $\RG_{-f}$ to obtain points in $\Dg_0(\RG_{-f})$ by pairing up-forks with maxima. Let $x, y \in  \RG_{f}$ correspond to the global minimum and maximum of $\bar{f}$ respectively. We pair $x$ with $y$, giving rise to the point $(\bar{f}(x),\bar{f}(y))$ in $\ExDg_0(\RG_{f})$ (see \figref{fig:persistence-diagram-Reeb-graph}(c)).

We are interested in the persistent features which are born and die within the range of $\bar{f}$. In \figref{fig:persistence-diagram-Reeb-graph}(b), the point $(1,\infty) \in \Dg_0(\RG_{f})$ corresponds to the unique homology class born at the global minimum of $f$ and persists throughout the sub-level set filtration. However, the point $(1,12) \in \ExDg_0(\RG_{f})$ encodes both the minimum and maximum of $\bar{f}$ (see \figref{fig:persistence-diagram-Reeb-graph}(c)). Therefore, $\Dg_0(\RG_{f})$ is combined with $\ExDg_0(\RG_{f})$ to obtain  $\PD_0(\RG_{f})$. This persistence diagram consists of the points in $\Dg_0(\RG_{f})$ except the point with infinite persistence $(1,\infty)$, instead the point  $(1,12)$ in $\ExDg_0(\RG_{f})$ is included. Thus $\PD_0(\RG_f):=\Dg_0(\RG_f)\cup \ExDg_0(\RG_f)\setminus \{(1,\infty)\}$ (see \figref{fig:persistence-diagram-Reeb-graph}(d)).

To compute the points in the $1$st extended persistence diagram $\ExDg_1(\RG_{f})$, we pair essential down-forks with essential up-forks. Let $u$ be an essential down-fork of $\RG_{f}$. Of all the cycles born at $u$, let $\gamma$ be a cycle having the largest minimum value of $\bar{f}$. Let $v \in \gamma$ be the node corresponding to the minimum on $\gamma$. It can be seen that $v$ is an essential up-fork \cite{2006-Agarwal-ExtremeElevation}. $u$ is paired with $v$ giving rise to the point $(\bar{f}(u), \bar{f}(v))$ in $\ExDg_1(f)$ (see \figref{fig:persistence-diagram-Reeb-graph}(e)). In this paper, we compute the persistence diagrams $\PD_0$ and $\ExDg_1$ of the Reeb graphs in an MDRG.

Next, we provide a brief description of the Laplace Beltrami operator and show its utility in computing descriptors of a shape.

\subsection{Laplace Beltrami Operator}
Let $\cM$ be a Riemannian $2$-manifold in $\mathbb{R}^3$. For a twice-differentiable function $f : \cM \rightarrow \mathbb{R}$, the Laplace Beltrami (LB) operator $\Delta_{\cM}$ is defined as follows:
\begin{small}
\begin{equation}
\Delta_{\cM}f = div(grad\text{ }f)
\end{equation}
\end{small}
where grad $f$ is the gradient of $f$ and div is the divergence on the manifold \cite{1984-chavel-eigenvalues}. Let $\psi$ be a diffeormorphism from an open set $U \subset \cM$ to $\mathbb{R}^2$ with
\begin{small}
\begin{align*}
g_{ij}&= \langle \partial_i \psi, \partial_j \psi \rangle, G = (g_{ij}),\\
W &= \sqrt{|G|}, (g^{ij})=G^{-1},
\end{align*}
\end{small}
where the matrix $G$ is a \textit{Riemannian tensor} on $\cM$ and $|G|$ is the determinant of $G$. The LB operator can now be written as follows \cite{1997-Rosenberg-Laplacian}:
\begin{small}
\begin{equation}
    \Delta_{\cM} f = \frac{1}{W} \sum_{i,j=1}^2 \partial_i ( g^{ij} W \partial_j f).
\end{equation}
\end{small}
The Riemannian metric $g$ determines the intrinsic properties of $\cM$, which are independent of the embedding of $\cM$. Further, since the definition of $g$ is based on inner products which are rotation invariant, $g$ is also invariant to rotations of $\cM$.

\subsubsection{Discretization of the Laplace Beltrami operator}
\begin{figure}
    \centering
    \includegraphics[width=5cm]{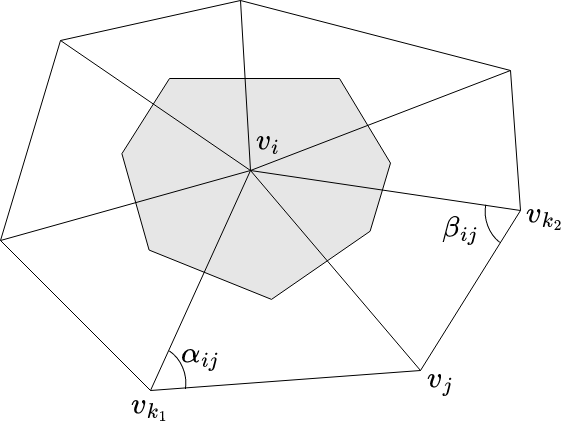}
    \caption{Definition of the angles $\alpha_{ij}$ and $\beta_{ij}$ and the area of the Voronoi cell (shaded region)  corresponding to vertex $v_i$}
    \label{fig:Cotangent-Weight-Sheme}
\end{figure}

Let $\M$ be a triangulation of the surface $\cM$, with $\mathcal{V} = \{v_1,v_2,...,v_n\}$ as the set of vertices. Two vertices in $\M$ are said to be adjacent if they are connected by an edge. The set of adjacent vertices of $v_i$ is denoted by $N(v_i)$. The LB operator at a vertex $v_i$ is written as follows:
\begin{small}
\begin{equation}
    \Delta_{\mathbb{M}} =\frac{1}{s_i} \sum_{v_j \in N(v_i)} \frac{\cot \alpha_{ij} + \cot \beta_{ij}}{2} [f(v_j) - f(v_i)].
\end{equation}
\end{small}
where, $s_i$ is the area of the Voronoi cell (shaded region) shown in \figref{fig:Cotangent-Weight-Sheme}. 

A weight function is defined as follows:
\begin{small}
\begin{equation}
    m_{ij} = \begin{cases}
      \frac{\cot \alpha_{ij} + \cot \beta_{ij}}{2} & \text{if $v_i$ and $v_j$ are adjacent}\\
      0 & \text{otherwise.}
    \end{cases}
\end{equation}
\end{small}
When $f$ is restricted to the vertices of $\M$, its discrete version is $\vec{f}$ with $f_i = f(v_i)$. The discrete LB operator of $f$ is written as $\Delta_{\M}f \approx L\vec{f}$. The entries of the matrix $L$ are given by
\begin{small}
\begin{equation*}
    L_{ij} = \begin{cases}
    \sum_k \frac{m_{ik}}{s_i}\text{ if $i = j$},\\
    -\frac{m_{ij}}{s_i}\text{ if $v_i$ and $v_j$ are adjacent},\\
    0\text{ otherwise.}
    \end{cases}
\end{equation*}
\end{small}
 The eigenvalues and eigenfunctions of the discrete version of the LB operator are computed by solving the eigenvalue problem $L\vec{v} = \lambda\vec{v}$. We note, the matrix $L$ is not symmetric since $s_i$ may not be the same for all vertices of $\M$. Therefore, the set of eigenvalues of this problem is not guaranteed to be real \cite{2007-Rustamov-Laplace-Beltrami-Eigenfunctions}. However, the matrix $L$ can be factorized as $L = S^{-1}M$, where $S$ is a diagonal matrix with $S_{ii} = s_i$ and $M$ is defined as follows:
 \begin{small}
\begin{equation}
    M_{ij} = \begin{cases}
    \sum_{i} m_{ij} \text{ if $i = j$},\\
    - m_{ij}\text{ if $v_i$ and $v_j$ are adjacent},\\
        0\text{ otherwise.}
    \end{cases}
\end{equation}
\end{small}
The eigenvalue problem $L\vec{v} = \lambda\vec{v}$ is rewritten as the generalized eigenvalue problem $S^{-1}M\vec{v} = \lambda \vec{v}$, or
\begin{small}
\begin{equation}
M\vec{v} = \lambda S \vec{v}.
\label{eqn:GeneralizedEigenvalueProblem}
\end{equation}
\end{small}
Since the matrices $S$ and $M$ are symmetric and $S$ is positive-definite, the eigenvalues and eigenvectors are real. Further, the eigenvalues are non-negative and the eigenvectors are orthogonal in terms of $S$-inner product:
\begin{small}
\begin{equation*}
    \langle \vec{u}, \vec{v}\rangle_{S} = \vec{u}^T S \vec{v}.
\end{equation*}
\end{small}
In this paper, we obtain the feature descriptors of a shape based on the eigenfunctions of LB operator.

\section{Proposed Distance between MDRGs}
\label{section:distance-measure-between-mdrgs}
In this section, we propose a distance measure between shapes by defining a distance measure between two MDRGs based on the bottleneck distance between Reeb graphs \cite{2014-Bauer-DistanceBetweenReebGraphs}. First we discuss our measure for bivariate fields.

\subsection{Distance between Bivariate Fields}
Let $\f = (f_1 , f_2 ),\, \g =(g_1, g_2)$ be two piecewise linear bivariate fields defined on $\M$ and let $\MDRG_{\f}$ and $\MDRG_{\g}$ be the corresponding MDRGs. To compute the proposed distance between $\MDRG_{\f}$ and $\MDRG_{\g}$, we compute a sum of  bottleneck distances between the component Reeb graphs in each dimension of the MDRGs by considering all possible bijections between the component Reeb graphs.

In the first dimension of the MDRGs, there is only one Reeb graph corresponding to each of the fields $f_1$ and $g_1$, denoted by $\RG_{f_1}$ and $\RG_{g_1}$, respectively. Let, $Range(f_1) = [\underset{x\in\M}{\min}f_1(x), \underset{x\in\M}{\max} f_1(x)]$ and $Range(g_1) = [\underset{x\in\M}{\min}g_1(x), \underset{x\in\M}{\max} g_1(x)]$ be the ranges of the functions $f_1$ and $g_1$. We define $R(f_1,g_1)=Range(f_1) \cup Range(g_1)$ as the union of the ranges of $f_1$ and $g_1$. Now for each $c \in R(f_1,g_1)$, in the second dimension, the  MDRGs $\MDRG_{\f}$ and $\MDRG_{\g}$ may have multiple Reeb graphs corresponding to $f_2$ and $g_2$ restricted on $f_1^{-1}(c)$ and $g_1^{-1}(c)$, respectively.
Let $\cS_{f_1}^{c} = \{ \RG_{\widetilde{f_2^p}} | \bar{f_1}(p) = c\}$ be the set of Reeb graphs from the second dimension of $\MDRG_{\f}$ and $\cS_{g_1}^{c} = \{ \RG_{\widetilde{g_2^p}} | \bar{g_1}(p) = c\}$ be the set of Reeb graphs from the second dimension of  $\MDRG_{\g}$. We consider all possible  bijections $\psi_c$ between $\cS_{f_1}^{c}$ and $\cS_{g_1}^{c}$. Then we define a distance between $\MDRG_{\f}$ and $\MDRG_{\g}$ as follows.
\begin{small}
\begin{multline}
    d_{\MDRG}\left(\MDRG_{\f}, \MDRG_{\g}\right) = d_B\left(\Dg\left(\RG_{f_1}\right), \Dg\left(\RG_{g_1}\right)\right) +\\ \frac{1}{|R(f_1,g_1)|}\underset{c\in R(f_1,g_1)}{\int} \underset{\psi_c : \cS_{f_1}^{c} \rightarrow \cS_{g_1}^{c}}{\inf} \sup_{\RG \in \cS_{f_1}^{c}} d_B\bigl(\Dg\left(\RG\right),\\ \hspace{6cm}\Dg\left(\psi_c\left(\RG\right)\right)\bigr)\\
   \label{eqn:bottleneck-distance-mdrgs}
\end{multline}
\end{small}
where $\Dg(\RG)$ is a persistence diagram of the Reeb graph $\RG$, $\psi_c$ ranges over bijections between $\cS_{f_1}^{c}$ and $\cS_{g_1}^{c}$. If $Range(f_1)\cap Range(g_1)\neq \emptyset$, then $|R(f_1,g_1)|$ is the length of the interval $R(f_1,g_1)$, otherwise $|R(f_1,g_1)|=|Range(f_1)|+|Range(g_1)|$.

In the current paper, corresponding to each of the Reeb graph in an MDRG, we construct two persistence diagrams, namely $\PD_0$ and $\ExDg_1$, as discussed in \secref{sec:persistence-diagram-of-reeb-graph}. We compute the distance between MDRGs defined in equation (\ref{eqn:bottleneck-distance-mdrgs}) based on $\PD_0$ and $\ExDg_1$, which are denoted by $d_{\MDRG}^{0}(\MDRG_{\f}, \MDRG_{\g})$ and $d_{\MDRG}^{1}(\MDRG_{\f}, \MDRG_{\g})$, respectively. We consider the proposed distance measure between $\MDRG_{\f}$ and $\MDRG_{\g}$ as the weighted sum of $d_{\MDRG}^{0}\left(\MDRG_{\f}, \MDRG_{\g}\right)$, $d_{\MDRG}^{0}\left(\MDRG_{-\f}, \MDRG_{-\g}\right)$ and $d_{\MDRG}^{1}\left(\MDRG_{\f}, \MDRG_{\g}\right)$.
\begin{small}
\begin{multline}
    d_{T}(\MDRG_{\f}, \MDRG_{\g}) = w_0 \cdot d_{\MDRG}^{0}\left(\MDRG_{\f}, \MDRG_{\g}\right)\\
    + w_1 \cdot d_{\MDRG}^{0}\left(\MDRG_{-\f}, \MDRG_{-\g}\right)
    + w_2 \cdot d_{\MDRG}^{1}\left(\MDRG_{\f}, \MDRG_{\g}\right)
\end{multline}
\end{small}
where, $0\leq w_0, w_1, w_2 \leq 1$ and $w_0 + w_1 + w_2= 1$. Here, $d_{\MDRG}^{0}\left(\MDRG_{\f}, \MDRG_{\g}\right)$ and $d_{\MDRG}^{0}\left(\MDRG_{-\f}, \MDRG_{-\g}\right)$ respectively compute the distance between persistence diagrams of sub-level set and super-level set filtrations of the Reeb graphs in the MDRGs and  $d_{\MDRG}^{1}\left(\MDRG_{\f}, \MDRG_{\g}\right)$ is the distance between the $1$-st extended persistence diagrams of the Reeb graphs in $\MDRG_{\f}$ and $\MDRG_{\g}$ respectively. Next, we discuss how to compute the proposed distance measure for quantized fields.
\subsubsection{Computational Aspects for Quantized Fields}
\label{subsubsec:Computational-Aspects}
We note, the MDRG of a bivariate field can be computed from the corresponding JCN, which is a quantized approximation of the Reeb space. For constructing the JCN of a bivariate field, the range of each of the component scalar fields is quantized or subdivided into a finite number of slabs. 
Algorithm \ref{algo:ComputePD} outlines the computation of the persistence diagrams by computing an MDRG corresponding to a bivariate field $\f=(f_1, f_2)$ and quantization levels $q_1$ and $q_2$. {\sc ConstructJCN} (in line 2) computes the JCN of $\f$ corresponding to $q_1$ and $q_2$. {\sc ConstructMDRG} (in line $3$) constructs the MDRG from the JCN. Finally, {\sc ComputePDReebGraph} computes the persistence diagram corresponding to a component Reeb graph (lines $6$-$8$). 

\begin{algorithm}
\caption{\sc{ComputePDsMDRG}}
\label{algo:ComputePD}
\textbf{Input:} $\f=(f_1,f_2), \; q_1, \; q_2$\\
\textbf{Output:} Persistence diagrams of the component Reeb graphs in the MDRG
\begin{algorithmic}[1]
\STATE \% \textit{Compute the JCN and MDRG of the bivariate field}
\STATE $\JCN_{\f}$ = {\sc ConstructJCN}$(\f, q_1, q_2)$
\STATE $\MDRG_{\f}$ = {\sc ConstructMDRG}$(\JCN_{\f})$
\STATE \% \textit{Compute the persistence diagrams of the component Reeb graphs in the MDRG}
\STATE $\Dg(\RG_{f_1})$ = {\sc ComputePDReebGraph}($\RG_{f_1}$)
\FOR{$p \in \RG_{f_1}$}
	\STATE $\Dg(\RG_{\widetilde{f_2^p}})$ = {\sc ComputePDReebGraph}($\RG_{\widetilde{f_2^p}}$)
\ENDFOR
\RETURN{$\left\{\Dg(\RG_{f_1}) , \left\{\Dg(\RG_{\widetilde{f_2^p}}) \middle| p \in \RG_{f_1}\right\}\right\}$}
\end{algorithmic}
\end{algorithm}

Algorithm \ref{algo:ComputeDistance} gives the pseudo-code for computing the distance between two multi-fields using the proposed distance measure. Let $\f=(f_1,g_2)$ and $\g = (g_1, g_2)$ be two bivariate fields defined on a compact $d$-manifold $\M$ with $d \geq 2$. Let $q_{1}$ and $q_{2}$ be the number of slabs (quantization levels) into which $\Range(f_1,g_1)$ and $\Range(f_2,g_2)$ are subdivided respectively, for constructing the JCNs of $\f$ and $\g$. Then let $c_1, c_2,\ldots, c_{q_1}$ be the quantized range values corresponding to the subdivision of $\Range(f_1,g_1)$. The distance between MDRGs defined in equation (\ref{eqn:bottleneck-distance-mdrgs}) can be written as:
\begin{small}
\begin{multline}
d_{\MDRG}\left(\MDRG_{\f}, \MDRG_{\g}\right) = d_B\left(\Dg\left(\RG_{f_1}\right), \Dg\left(\RG_{g_1}\right)\right) +\\ \frac{1}{q_1}\underset{1 \leq i \leq q_1}{\sum} \underset{\psi_{c_{i}} : \cS_{f_1}^{c_i} \rightarrow \cS_{g_1}^{c_i}}{\inf} \sup_{\RG \in \cS_{f_1}^{c_i}} d_B\left(\Dg\left(\RG\right), \Dg\left(\psi_{c_i}\left(\RG\right)\right)\right)
   \label{eqn:bottleneck-distance-mdrgs-quantized}
\end{multline}
\end{small}
\noindent
where $\psi_{c_i}$ ranges over all possible  bijections between $\cS_{f_1}^{c_i}$ and $\cS_{g_1}^{c_i}$. Since the number of Reeb graphs in $\cS_{f_1}^{c_i}$ and $\cS_{g_1}^{c_i}$ need not be equal, it may not be possible to construct proper bijections between $\cS_{f_1}^{c_i}$ and $\cS_{g_1}^{c_i}$. Therefore, we introduce dummy Reeb graphs (without vertices and edges) in $\cS_{f_1}^{c_i}$ or $\cS_{g_1}^{c_i}$ to make the cardinality of $\cS_{f_1}^{c}$ and $\cS_{g_1}^{c_i}$ equal. Then the optimal bijection $\psi_{c_i}$ is constructed using the Hungarian algorithm by computing a cost matrix \cite{1955-Kuhn-Hungarian-Algorithm} (lines $8$-$12$ of Algorithm \ref{algo:ComputeDistance}). We note, the bijection $\psi_{c_i}$ can map a Reeb graph in $\cS_{f_1}^{c_i}$ to a dummy Reeb graph. The persistence diagram corresponding to a dummy Reeb graph is empty, i.e, it does not contain any point. The bottleneck distance between persistence diagrams is computed as described in \secref{subsec:bottleneck-distance-between-persistence-diagrams}.

\begin{algorithm}
\caption{\sc{ComputeDistance}}
\label{algo:ComputeDistance}
\textbf{Input:} $\f,\; \g,\; q_1, \; q_2$\\
\textbf{Output:} $d_{\MDRG}(\MDRG_{\f}, \MDRG_{\g})$
\begin{algorithmic}[1]
\STATE \% \textit{Compute the Persistence Diagrams}
\STATE $PD_{\f}$ = {\sc ComputePDsMDRG}($\f, q_1, q_2$)
\STATE $PD_{\g}$ = {\sc ComputePDsMDRG}($\g, q_1, q_2$)
\STATE \% \textit{Compute the distance between MDRGs}
\STATE $Range(f_1)$ = $[\underset{x\in\M}{\min}f_1(x), \underset{x\in\M}{\max} f_1(x)]$
\STATE $Range(g_1)$ = $[\underset{x\in\M}{\min}g_1(x), \underset{x\in\M}{\max} g_1(x)]$
\STATE $R(f_1,g_1)$ = $Range(f_1) \cup Range(g_1)$
\FOR{$i = 1,2,\ldots,q_1$}
	\STATE $\cS_{f_1}^{c_i}$ = $\{ \RG_{\widetilde{f_2^p}} | \bar{f_1}(p) = c_i\}$
	\STATE $\cS_{g_1}^{c_i}$ = $\{ \RG_{\widetilde{g_2^p}} | \bar{g_1}(p) = c_i\}$
        \STATE Compute optimal bijection $\psi_{c_i}$ between elements of $\cS_{f_1}^{c_i}$ and $\cS_{g_1}^{c_i}$ (by computing the bottleneck distances between the corresponding persistence diagrams in $PD_{\f}$ and $PD_{\g}$), using the Hungarian algorithm
\ENDFOR
\STATE \begin{small}$\displaystyle d_{\MDRG}\left(\MDRG_{\f}, \MDRG_{\g}\right)$ = $d_B\left(\Dg\left(\RG_{f_1}\right), \Dg\left(\RG_{g_1}\right)\right) +$\\
$\hspace{1.4cm}\displaystyle\frac{1}{q_1}\underset{1 \leq i \leq q_1}{\sum} \sup_{\RG \in \cS_{f_1}^{c_i}} d_B\left(\Dg\left(\RG\right), \Dg\left(\psi_{c_i}\left(\RG\right)\right)\right)$\end{small}
\RETURN{$ d_{\MDRG}\left(\MDRG_{\f}, \MDRG_{\g}\right)$}
\end{algorithmic}
\end{algorithm}
\subsection{Properties of the Proposed Distance Measure}
\label{Distance-Measure-Properties}
In this sub-section, we discuss the properties of the proposed distance between MDRGs. Let $\f=(f_1, f_2), \g=(g_1,g_2)$ be two bivariate fields. First, we show that $d_{T}(\MDRG_{\f},\MDRG_{\g})$ is a pseudo-metric.
\begin{theorem}
$d_{T}$ satisfies the properties of a pseudo-metric.
\label{thm:psuedo-metric}
\end{theorem}
\begin{proof}
\textbf{Non-negativity:} For two MDRGs $\MDRG_{\f}$ and $\MDRG_{\g}$, we have $d_{T}(\MDRG_{\f}, \MDRG_{\g}) \geq 0$. Thus the non-negativity property is satisfied.

\noindent
\textbf{Symmetry:} For two MDRGs $\MDRG_{\f}$ and $\MDRG_{\g}$, we have $d_{T}(\MDRG_{\f}, \MDRG_{\g}) = d_{T}(\MDRG_{\g}, \MDRG_{\f})$. Thus the symmetry property holds.

\noindent
\textbf{Identity:} Bauer $\etal$ \cite{2014-Bauer-DistanceBetweenReebGraphs} showed that two distinct Reeb graphs can have the same persistence diagram and therefore the bottleneck distance between Reeb graphs is a pseudo-metric. Since we compute $d_{T}(\MDRG_{\f}, \MDRG_{\g})$ by calculating the bottleneck distance between the component Reeb graphs of the MDRGs $\MDRG_{\f}$ and $\MDRG_{\g}$, it follows that the bottleneck distance between two non-identical MDRGs $\MDRG_{\f}$ and $\MDRG_{\g}$ can be zero. Therefore, the identity property does not hold.

\noindent
\textbf{Triangle inequality:} Let us consider three MDRGs $\MDRG_{\f}, \MDRG_{\g}$ and $\MDRG_{\h}$, corresponding to bivariate fields $\f=(f_1,f_2), \g=(g_1,g_2)$ and $\h=(h_1,h_2)$, respectively. We first prove the triangle inequality property for $d_{\MDRG}^{0}$ between MDRGs. We need to show that $d_{\MDRG}^{0}(\MDRG_{\f}, \MDRG_{\h}) \leq d_{\MDRG}^{0}(\MDRG_{\f}, \MDRG_{\g}) + d_{\MDRG}^{0}(\MDRG_{\g}, \MDRG_{\h})$.

\noindent
For $c \in R(f_1,g_1)$, let $\psi_c'$ be the bijection between $\cS_{f_1}^{c}$ and $\cS_{g_1}^{c}$ achieving $d_{\MDRG}^{0}(\MDRG_{\f},\MDRG_{\g})$. Similarly, let $\psi_c''$ be the bijection between $\cS_{g_1}^{c}$ and $\cS_{h_1}^{c}$ achieving $d_{\MDRG}^{0}(\MDRG_{\g},\MDRG_{\h})$. Let $\psi_c'''$ be the bijection between $\cS_{f_1}^{c}$ and $\cS_{h_1}^{c}$ obtained by the composition of $\psi_c''$ and $\psi_c'$, i.e, $\psi_c''' = \psi_c'' \circ \psi_c'$.
\begin{small}
\begin{align*}
&d_{\MDRG}^{0}\left(\MDRG_{\f}, \MDRG_{\h}\right)\\
&= d_B\left(\PD_0\left(\RG_{f_1}\right),\PD_0\left(\RG_{h_1}\right)\right) +\\
&\frac{1}{|R(f_1,h_1)|}\underset{c\in R(f_1,h_1)}{\int}\underset{\psi_c : \cS_{f_1}^{c} \rightarrow \cS_{h_1}^{c}}{\inf} \sup_{\RG \in \cS_{f_1}^{c}} d_B\bigl(\PD_0\left(\RG\right),\\
&\hspace{6cm}\PD_0\left(\psi_c\left(\RG\right)\right)\bigr)\\
&\leq d_B\left(\PD_0\left(\RG_{f_1}\right), \PD_0\left(\RG_{h_1}\right)\right) +\\
&\hspace{0.4cm}\frac{1}{|R(f_1,h_1)|}\underset{c\in R(f_1,h_1)}{\int} \sup_{\RG \in \cS_{f_1}^{c}} d_B\bigl(\PD_0\left(\RG\right),\\
&\hspace{5cm}\PD_0\left(\psi_c'''\left(\RG\right)\right)\bigr)\\
&\leq d_B\left(\PD_0\left(\RG_{f_1}\right), \PD_0\left(\RG_{g_1}\right)\right) +  d_B\bigl(\PD_0\left(\RG_{g_1}\right),\\
&\hspace{5.7cm}\PD_0\left(\RG_{h_1}\right)\bigr) +\\\\\\
&\hspace{0.4cm}\frac{1}{|R(f_1,h_1)|}\underset{c\in R(f_1,h_1)}{\int} \sup_{\RG \in \cS_{f_1}^{c}} \Bigg(d_B\bigl(\PD_0\left(\RG\right),\\
&\hspace{5cm}\PD_0\left(\psi_c'\left(\RG\right)\right)\bigr) +\\
&\hspace{2cm}d_B\left(\psi_c'\left(\PD_0\left(\RG\right)\right), \PD_0\left(\psi_c''\circ \psi_c'\left(\RG\right)\right)\right)\Bigg)\\
&\hspace{1cm}\text{(since $d_B$ between Reeb graphs is a pseudo-metric)}\\
&\leq d_B\left(\PD_0\left(\RG_{f_1}\right), \PD_0\left(\RG_{g_1}\right)\right) +  d_B\bigl(\PD_0\left(\RG_{g_1}\right),\\
&\hspace{5.7cm}\PD_0\left(\RG_{h_1}\right)\bigr) +\\
&\hspace{0.4cm}\frac{1}{|R(f_1,g_1)|}\underset{c\in R(f_1,g_1)}{\int} \sup_{\RG \in \cS_{f_1}^{c}} d_B\bigl(\PD_0\left(\RG\right),\\
&\hspace{5cm}\PD_0\left(\psi_c'\left(\RG\right)\right)\bigr) +\\
&\hspace{0.4cm}\frac{1}{|R(g_1,h_1)|}\underset{c\in R(g_1,h_1)}{\int} \sup_{\RG \in \cS_{g_1}^{c}} d_B\bigl(\PD_0\left(\RG\right),\\
&\hspace{5cm}\PD_0\left(\psi_c''\left(\RG\right)\right)\bigr)\\
&= d_{\MDRG}^{0}(\MDRG_{\f}, \MDRG_{\g}) + d_{\MDRG}^{0}(\MDRG_{\g}, \MDRG_{\h}).
\end{align*}
\end{small}

Using a similar argument, we can prove the triangle inequality for $d_{\MDRG}^{1}$. We now show the triangle inequality property for $d_{T}(\MDRG_{\f}, \MDRG_{\h})$.
\begin{small}
\begin{align*}
&d_{T}(\MDRG_{\f}, \MDRG_{\h})\\
&= w_0 \cdot d_{\MDRG}^{0}\left(\MDRG_{\f}, \MDRG_{\h}\right) + w_1 \cdot d_{\MDRG}^{0}\left(\MDRG_{-\f}, \MDRG_{-\h}\right)\\
&\hspace{3cm} + w_2 \cdot d_{\MDRG}^{1}\left(\MDRG_{\f}, \MDRG_{\h}\right)\\
&\leq w_0\left(d_{\MDRG}^{0}\left(\MDRG_{\f}, \MDRG_{\g}\right)+d_{\MDRG}^{0}\left(\MDRG_{\g}, \MDRG_{\h}\right)\right) +\\
&\hspace{0.4cm} w_1\left(d_{\MDRG}^{0}\left(\MDRG_{-\f}, \MDRG_{-\g}\right)+d_{\MDRG}^{0}\left(\MDRG_{-\g}, \MDRG_{-\h}\right)\right) +\\
&\hspace{0.4cm} w_2\left(d_{\MDRG}^{1}\left(\MDRG_{\f}, \MDRG_{\g}\right)+d_{\MDRG}^{1}\left(\MDRG_{\g}, \MDRG_{\h}\right)\right)\\
&= w_0 \cdot d_{\MDRG}^{0}\left(\MDRG_{\f}, \MDRG_{\g}\right) + w_1 \cdot d_{\MDRG}^{0}\left(\MDRG_{-\f}, \MDRG_{-\g}\right)\\
&\hspace{0.4cm} +  w_2 \cdot d_{\MDRG}^{1}\left(\MDRG_{\f}, \MDRG_{\g}\right) + w_0 \cdot d_{\MDRG}^{0}\left(\MDRG_{\g}, \MDRG_{\h}\right)\\
&\hspace{0.4cm} + w_1 \cdot d_{\MDRG}^{0}\left(\MDRG_{-\g}, \MDRG_{-\h}\right) + w_2 \cdot d_{\MDRG}^{1}\left(\MDRG_{\g}, \MDRG_{\h}\right)\\
&=d_{T}(\MDRG_{\f}, \MDRG_{\g}) + d_{T}(\MDRG_{\g}, \MDRG_{\h}).
\end{align*}
\end{small}
\end{proof}
Next, we prove the stability of the proposed measure $d_T$.  Let $\f = (f_1, f_2) : \M \rightarrow \mathbb{R}^2$ and  $\g = (g_1, g_2) : \M \rightarrow \mathbb{R}^2$ be two bivariate fields defined on a compact metric space $\M$ such that $f_1, f_2, g_1$ and $g_2$ are tame functions.
Let $\MDRG_{\f}$ and $\MDRG_{\g}$ be the MDRGs of $\f$ and $\g$, respectively. Then we prove the following lemma.

\begin{lemma}
\label{stability-lemma-1}
\begin{small}
\begin{equation*}
d_{\MDRG}^0(\MDRG_{\f},\MDRG_{\g})\leq \|f_1 - g_1\|_{\infty} + \frac{1}{2}\max \{Amp(f_2), Amp(g_2) \}
\end{equation*}
\end{small}
where $Amp(f)$ denotes the amplitude of a real-valued function $f:\M \rightarrow \R$ and is defined as follows:
\begin{small}
\begin{equation*}
    Amp(f) = \max_{x\in\M} f(x) -  \min_{y\in\M} f(y).
\end{equation*}
\end{small}
\end{lemma}
\begin{proof}
The term $d_B\left(\PD_0\left(\RG_{f_1}\right), \PD_0\left(\RG_{g_1}\right)\right)$ in $d_{\MDRG}^{0}(\MDRG_{\f},\MDRG_{\g})$ of equation ($10$) is bounded above by $\|f_1 - g_1\|_{\infty}$ (see \cite{2007-Cohen-Steiner-Bottleneck} for more details). We now prove a bound on the term $\underset{ c \in \Range(f_1, g_1)}{\int}\underset{\psi : \cS_{f_1}^{c} \rightarrow \cS_{g_1}^{c}}{\inf} \underset{\RG \in \cS_{f_1}^{c}}{\sup} d_B\left(\PD_0\left(\RG\right), \PD_0\left(\psi\left(\RG\right)\right)\right)$.

Let $\RG \in \cS_{f_1}^{c}$. Now, every point in the persistence diagram $\PD_0\left(\RG\right)$ corresponds to a $0$-dimensional class whose birth and death lies between $\underset{x \in \RG}{\min} \overline{\widetilde{f_2^p}}(x)$ and $\underset{x \in \RG}{\max}\overline{\widetilde{f_2^p}}(x)$. Therefore, the maximum persistence of a point in $\PD_0\left(\RG\right)$ is at most the amplitude of $\overline{\widetilde{f_2^p}}$ in $\RG$.
\begin{small}
\begin{align}
\nonumber \max_{x \in \PD_0\left(\RG\right)} pers(x) &\leq \max_{p \in \RG_{\widetilde{f_2^p}}} \overline{\widetilde{f_2^p}}(p) -  \min_{p \in \RG_{\widetilde{f_2^p}}} \overline{\widetilde{f_2^p}}(p).\\
\nonumber &\leq \max_{p \in \M} f_2(p) -  \min_{p \in \M} f_2(p).\\
&= Amp(f_2) \label{eqn:bound-persistence-f}.
\end{align}
\end{small}
Similarly, for $\RG' \in \cS_{g_2}^{c}$, the following bound is obtained on the maximum persistence of a point in the persistence diagram $\PD_0\left(\RG'\right)$.
\begin{small}
\begin{align}
\max_{x \in \PD_0\left(\RG'\right)} pers(x) &\leq Amp(g_2) \label{eqn:bound-persistence-g}.
\end{align}
\end{small}
Let $(a,b) \in \PD_0(\RG)$ and $(c,d)$ be its matching pair in $\PD_0(\RG')$ corresponding to $d_B\left(\PD_0\left(\RG\right), \PD_0\left(\RG'\right)\right)$. Then $\|(a,b) - (c,d)\|_{\infty}$ is at most $\max\{\frac{a+b}{2},\frac{c+d}{2}\}$. Otherwise, we can match $(a,b)$ to $(\frac{a+b}{2}, \frac{a+b}{2})$, $(c,d)$  to $(\frac{c+d}{2}, \frac{c+d}{2})$ and obtain the required bound. Therefore, for any matched pair of points $(x,x')$, $\|x - x'\|_{\infty}$ is bounded by $\max\{\frac{1}{2}pers(x), \frac{1}{2}pers(x')\}$. We now obtain a bound on $d_B\left(\PD_0\left(\RG\right), \PD_0\left(\RG'\right)\right)$ as follows:
\begin{small}
\begin{align*}
&d_B\left(\PD_0\left(\RG\right), \PD_0\left(\RG'\right)\right)\\
&\leq \max \left\{\max_{x \in \PD_0\left(\RG\right)} \frac{1}{2} pers(x),  \max_{x \in \PD_0\left(\RG'\right)} \frac{1}{2}pers(x)\right\}\\
&\leq \frac{1}{2} \max \{Amp(f_2), Amp(g_2)\}\\
&\hspace{1cm}(\text{from equations (\ref{eqn:bound-persistence-f}})\text{ and }(\ref{eqn:bound-persistence-g})).
\end{align*}
\begin{align*}
    \nonumber &d_{\MDRG}^0(\MDRG_{\f} , \MDRG_{\g})\\
    &= d_B\left(\PD_0\left(\RG_{f_1}\right), \PD_0\left(\RG_{g_1}\right)\right) +\\ 
    &\hspace{0.2cm}\frac{1}{|\Range(f_1, g_1)|}\underset{c\in\Range(f_1, g_1)}{\int} \underset{\psi : \cS_{f_1}^{c} \rightarrow \cS_{g_1}^{c}}{\inf} \sup_{\RG \in \cS_{f_1}^{c}} d_B\bigl(\PD_0\left(\RG\right),\\
    &\hspace{6.1cm}\PD_0\left(\psi\left(\RG\right)\right)\bigr)\\
    &\leq \| f_1 - g_1\|_{\infty} + \frac{1}{|\Range(f_1, g_1)|}\underset{c\in\Range(f_1, g_1)}{\int} \frac{1}{2}\max \{Amp(f_2),\\
    &\hspace{6.4cm}Amp(g_2)\}.\\
    &= \| f_1 - g_1\|_{\infty} + \frac{1}{2}\max \{Amp(f_2), Amp(g_2)\}.\\ 
\end{align*}
\end{small}
\end{proof}
\begin{lemma}
\label{stability-lemma-2}
\begin{small}
\begin{equation*}
d_{\MDRG}^1(\MDRG_{\f},\MDRG_{\g})\leq 3\|f_1 - g_1\|_{\infty} + \frac{1}{2}\max \{Amp(f_2), Amp(g_2) \}.
\end{equation*}
\end{small}
\end{lemma}
\begin{proof}
The term $d_B\left(\ExPD_0\left(\RG_{f_1}\right), \ExPD_0\left(\RG_{g_1}\right)\right)$ in $d_{\MDRG}^{1}(\MDRG_{\f}, \MDRG_{\g})$ is bounded above by $3\|f_1 - g_1\|_{\infty}$ (see Theorems $4.1$ and $4.3$ in \cite{2014-Bauer-DistanceBetweenReebGraphs} for more details). The rest of the proof is similar to that of Lemma \ref{stability-lemma-1}.
\end{proof}
The following theorem gives the stability of $ d_{T}(\MDRG_{\f},\MDRG_{\g})$.
\begin{theorem}
\label{theorem:stability}
\textbf{Stability: }
\begin{small}
\begin{equation*}
    d_{T}(\MDRG_{\f},\MDRG_{\g})\leq 3\|f_1 - g_1\|_{\infty} + \frac{1}{2}\max \{Amp(f_2), Amp(g_2) \}.
\end{equation*}
\end{small}
\end{theorem}
\begin{proof}
\begin{small}
\begin{align*}
&d_{T}(\MDRG_{\f},\MDRG_{\g})\\
&= w_0 \cdot d_{\MDRG}^{0}(\MDRG_{\f}, \MDRG_{\g}) + w_1 \cdot d_{\MDRG}^{0}(\MDRG_{-\f}, \MDRG_{-\g}) +\\
& \hspace{3.2cm} w_2 \cdot d_{\MDRG}^{1}(\MDRG_{\f}, \MDRG_{\g})\\
&\leq w_0\left(\|f_1 - g_1\|_{\infty} + \frac{1}{2}\max \{Amp(f_2), Amp(g_2)\}\right) +\\
&\hspace{0.4cm} w_1\left(\|f_1 - g_1\|_{\infty} + \frac{1}{2}\max \{Amp(f_2), Amp(g_2)\}\right) +\\
&\hspace{0.4cm}w_2\left(3\|f_1 - g_1\|_{\infty} + \frac{1}{2}\max \{Amp(f_2), Amp(g_2)\}\right)\\
&\hspace{3cm} \text{(using Lemmas \ref{stability-lemma-1}, \ref{stability-lemma-2})}\\
&\leq 3\|f_1 - g_1\|_{\infty} + \frac{1}{2}\max \{Amp(f_2), Amp(g_2)\}\\
&\hspace{3cm} (\text{since } w_0 + w_1 + w_2 = 1).
\end{align*}
\end{small}
\end{proof}

\subsection{Complexity Analysis}
\label{subsec:Complexity-Analysis}
In this sub-section, we analyze the time and space complexities of computing the distance between two bivariate fields based on the corresponding MDRGs. First, we give the complexity for constructing the MDRG of a bivariate field and computing the persistence diagrams of the component Reeb graphs of the MDRG. Then, we analyze the complexity of computing the proposed distance measure between two MDRGs.

\subsubsection{Algorithm \ref{algo:ComputePD}: Computing the MDRG and Persistence Diagrams of its Component Reeb Graphs}
\label{subsubsec:complexity-PD-Computation}
\textbf{Time Complexity.} Let $\f=(f_1,f_2)$ be a bivariate field defined on a compact $d$-manifold. To construct the MDRG of $\f$, denoted by $\MDRG_{\f}$, we require a JCN as input. The construction of the JCN takes $\cO(2|E_1| + |E_1|\alpha(|E_1|))$ time, where $|E_1|$ is the number of edges in the JCN and $\alpha$ is the inverse Ackermann function \cite{1975-Tarjan-UF} (line $2$, Algorithm \ref{algo:ComputePD}). The time complexity of constructing the MDRG from the JCN  using the algorithm in \cite{2016-Chattopadhyay-MultivariateTopologySimplification} is $\cO(2|V_1|(|V_1| + |E_1| \alpha(|V_1|) + |V_1| \log(|V_1|)))$, where $|V_1|$ is the number of vertices of the JCN (line $3$, Algorithm \ref{algo:ComputePD}). We note, the Reeb graph $\RG_{f_1}$ has at most $|V_1|$ vertices and $|E_1|$ edges. Therefore, $\RG_{f_1}$ has at most $|V_1| + |E_1|$ simplices (vertices and edges). The computation of the persistence diagram of $\RG_{f_1}$ take $\cO((|V_1| + |E_1|)^3)$ time \cite{2010-Edelsbrunner-book} (line $5$, Algorithm \ref{algo:ComputePD}). Next, we see the time complexity of computing the persistence diagrams of the Reeb graphs of $f_2$ in $\MDRG_{\f}$ (lines $6$-$8$, Algorithm \ref{algo:ComputePD}). 

We note, each node/edge in a Reeb graph $\RG_{\widetilde{f_2^p}}$ ($p \in \RG_{f_1}$) corresponds to a unique node/edge in the JCN \cite{2016-Chattopadhyay-MultivariateTopologySimplification}. Therefore, the total number of simplices (vertices and edges) in the second-dimensional Reeb graphs of $\MDRG_{\f}$ is at most the total number of simplices in the JCN. Thus, the construction of the persistence diagrams of the Reeb graphs of $f_2$, i.e. $\RG_{\widetilde{f_2^p}} (p \in \RG_{f_1})$ take $\cO((|V| + |E|)^3)$ time. The total time for constructing the MDRG of $\f$ and computing the persistence diagrams of its component Reeb graphs (in Algorithm \ref{algo:ComputePD}) is $\cO(2|V_1|(|V_1| + |E_1| \alpha(|V_1|) + |V_1| \log(|V_1|))) + \cO((|V_1| + |E_1|)^3) + \cO((|V_1| + |E_1|)^3) = \cO(2(|V_1| + |E_1|)^3)$. Next, we analyze the space complexity of computing $\MDRG_{\f}$ and persistence diagrams of the Reeb graphs in $\MDRG_{\f}$.

\textbf{Space Complexity.} The space complexity of constructing the JCN of $\f$ depends on the number of edges it contains $(|E_1|)$ and the number of fragments in the domain corresponding to the quantization of $\f$, denoted by $N_{\f}$. Since $|E_1| = \cO((4 + d)N_{\f})$ \cite{2014-Carr-JCN}, the space complexity of computing the JCN is $\cO(|E_1|)$. The number of nodes/edges in $\RG_{f_1}$ is bounded by the number of nodes/edges in the JCN. Thus the Reeb graph $\RG_{f_1}$ takes $\cO(|V_1| +|E_1|)$ space. Similarly, based on the bound obtained on the number of simplices in the Reeb graphs $\RG_{\widetilde{f_2^p}} (p \in \RG_{f_1})$, the Reeb graphs in the second dimension of $\MDRG_{\f}$ take $\cO(|V_1| + |E_1|)$ space. Thus, the space complexity of $\MDRG_{\f}$ is $\cO(2(|V_1|+|E_1|))$. Next, we analyze the space complexity for the persistence diagrams of the Reeb graphs in $\MDRG_{\f}$. We note, the persistence diagram of a Reeb graph is obtained by pairing its critical nodes. Therefore, the number of points in a persistence diagram is at most half the number of vertices in the corresponding Reeb graph. The total number of nodes in the component Reeb graphs of $\MDRG_{\f}$ is at most $2|V_1|$. Thus, the total number of points in the persistence diagrams of the Reeb graphs in $\MDRG_{\f}$ is $|V_1|$. The total space complexity of computing $\MDRG_{\f}$ and the persistence diagrams of its component Reeb graphs is $\cO(2(|V_1| + |E_1|) + |V_1|) = \cO(3|V_1| + 2|E_1|)$.

\subsubsection{Algorithm \ref{algo:ComputeDistance}: Computing the Distance between MDRGs}
\textbf{Time Complexity.} Let $\f=(f_1, f_2), \g=(g_1, g_2)$ be two bivariate fields and $\MDRG_{\f}, \MDRG_{\g}$ be the corresponding MDRGs. Let $(|V_1|, |E_1|)$ and $(|V_2|,|E_2|)$ be the number of vertices and edges in the JCNs of $\f$ and $\g$ respectively. Then the total number of points in the persistence diagrams of the component Reeb graphs of $\MDRG_{\f}$ and $\MDRG_{\g}$ are at most $|V_1|$ and $|V_2|$ respectively (see \secref{subsubsec:complexity-PD-Computation} for more details). Therefore, the total time for computing $d_B\left(\Dg\left(\RG_{f_1}\right), \Dg\left(\RG_{g_1}\right)\right)$ and $d_B\left(\Dg\left(\RG_{\widetilde{f_2^p}}\right), \Dg\left(\RG_{\widetilde{g_2^{p'}}}\right)\right)$ $\forall p \in \RG_{f_1}, p' \in \RG_{g_1}$ using the Hungarian algorithm is $\cO((|V_1| + |V_2|)^{3})$ \cite{1955-Kuhn-Hungarian-Algorithm}. 

Let $q_{1}$ be the number of slabs into which $\Range(f_1,g_1)$ is subdivided, for constructing the JCNs of $\f$ and $\g$. Then let $c_1, c_2,\ldots, c_{q_1}$ be the quantized range values corresponding to the subdivision of $\Range(f_1,g_1)$. For each $c_i$, we need to compute a map $\psi_{c_{i}} : \cS_{f_1}^{c_i} \rightarrow \cS_{g_1}^{c_i}$ which achieves the minimal value of $\sup_{\RG \in \cS_{f_1}^{c_i}} d_B\left(\Dg\left(\RG\right),\Dg(\psi_{c_i}(\RG))\right)$ (lines $8$-$12$, Algorithm \ref{algo:ComputeDistance}). To do so, we need to compute $d_B\left(\Dg\left(\RG\right), \Dg\left(\RG'\right)\right)$ for each $\Dg \in \cS_{f_1}^{c_i}$ and $\Dg' \in \cS_{g_1}^{c_i}$. The total time for computing $\displaystyle \bigcup_{1\leq i\leq q_1} \bigcup_{\RG \in \cS_{f_1}^{c_i}, \RG' \in \cS_{g_1}^{c_i}} d_B\left(\Dg\left(\RG\right), \Dg\left(\RG'\right)\right)$ is bounded by the time complexity for computing the bottleneck distance between all pairs of Reeb graphs in the second dimension of $\MDRG_{\f}$ and $\MDRG_{\g}$, which is $\cO((|V_1| + |V_2|)^{3})$. After this step, the computation of all the optimal bijections $\psi_{c_i} (i \in \{1,2,\ldots,q_1\})$ using Hungarian algorithm take $\cO((|V_1| + |V_2|)^3)$ time \cite{1955-Kuhn-Hungarian-Algorithm}. The total time for computing the distance between the collections of persistence diagrams of $\MDRG_\f$ and $\MDRG_\g$ (in Algorithm \ref{algo:ComputeDistance}) is $\cO((|V_1| + |V_2|)^{3}) + \cO((|V_1| + |V_2|)^3) = \cO(2(|V_1| + |V_2|)^3)$. Next, we analyze the space complexity of the proposed distance between MDRGs. 

\textbf{Space Complexity.} The distance between $\MDRG_{\f}$ and $\MDRG_{\g}$ is computed based on the bottleneck distance between the persistence diagrams of the component Reeb graphs in the respective MDRGs. To compute the bottleneck distance between the persistence diagrams of two Reeb graphs, we construct an optimal bijection between the points in the persistence diagrams using the Hungarian algorithm. For two persistence diagrams consisting of $N_1$ and $N_2$ points respectively, the Hungarian algorithm takes $\cO((N_1 + N_2)^2)$ space. Based on the bound obtained on the number of points in the persistence diagrams of $\RG_{f_1}$ and $\RG_{g_1}$, the space complexity for computing $d_B\left(\Dg\left(\RG_{f_1}\right), \Dg\left(\RG_{g_1}\right)\right)$ is $\cO((\frac{1}{2}|V_1| + \frac{1}{2}|V_2|)^2)$. Similarly, the total space complexity for the computation of $d_B\left(\Dg\left(\RG_{\widetilde{f_2^p}}\right), \Dg\left(\RG_{\widetilde{g_2^{p'}}}\right)\right)$ for all combinations of $p \in \RG_{f_1}$ and $p' \in \RG_{g_1}$, is $\cO((\frac{1}{2}|V_1| + \frac{1}{2}|V_2|)^2)$. Thus the computation of distance between MDRGs takes $\cO(\frac{1}{2}(|V_1|^2+|V_2|^2) + |V_1||V_2|)$ space.

\subsection{Generalization for more than two fields}
\label{subsec:generalizatio-for-more-than-two-fields}
Although, in this paper, we deal with MDRGs of bivariate fields, the proposed distance measure can be extended for MDRGs corresponding to more than $2$ fields. If $\f=(f_1,f_2,..,f_r) : \M \rightarrow \R^r$ and $\g=(g_1,g_2,...,g_r) : \M \rightarrow \R^r$ are multi-fields defined on a triangulated mesh $\M$ of dimension $d \geq r$, then the distance between the MDRGs $\MDRG_{\f}$ and $\MDRG_{\g}$ is defined as follows:
\begin{small}
\begin{multline}
d_{\MDRG,r}^k\left(\MDRG_{\f}, \MDRG_{\g}\right) = d_B\left(\Dg_k\left(\RG_{f_1}\right), \Dg_k\left(\RG_{g_1}\right)\right) +\\ \sum_{i=1}^{r-1}\frac{1}{|R(f_i,g_i)|}\underset{c\in R(f_i, g_i)}{\int}\underset{\psi^{(i)}_c : \cS_{f_i}^{c} \rightarrow \cS_{g_i}^{c}}{\inf}
\sup_{\RG \in \cS_{f_{i}}^{c}} d_B\bigl(\Dg_k(\RG),\\
\hspace{6.1cm}\Dg_k(\psi_i(\RG))\bigr).
\label{eqn:generalization-bottleneck-distance-mdrgs}
\end{multline}
\end{small}
Here, $\cS_{f_i}^{c}$ and $\cS_{g_i}^{c}$ denote the collections of Reeb graphs in the $(i+1)^{th}$ dimension of $\MDRG_{\f}$ and $\MDRG_{\g}$ respectively, $\psi^{(i)}_c$ varies over bijections between $\cS_{f_i}^{c}$ and $\cS_{g_i}^{c}$, and $R(f_i,g_i)$ represents the union of the ranges of $f_i$ and $g_i$. Similar to the case of bivariate fields, we denote the distance between MDRGs based on the persistence diagrams $\PD_0$ and $\ExDg_1$ by $d_{\MDRG, r}^{0}(\MDRG_{\f}, \MDRG_{\g})$ and $d_{\MDRG, r}^{1}(\MDRG_{\f}, \MDRG_{\g})$, respectively. We now define a distance between $\MDRG_{\f}$ and $\MDRG_{\g}$ as follows.
\begin{small}
\begin{multline}
    d_{T,r}(\MDRG_{\f},\MDRG_{\g}) = d_{\MDRG,r}^{0}(\MDRG_{\f},\MDRG_{\g})\\
    + d_{\MDRG,r}^{0}(\MDRG_{-\f},\MDRG_{-\g}) +d_{\MDRG,r}^{1}(\MDRG_{\f},\MDRG_{\g}) 
\end{multline}
\end{small}
where, $0\leq w_0, w_1, w_2 \leq 1$ and $w_0 + w_1 + w_2= 1$. Here, $d_{\MDRG,r}^{0}\left(\MDRG_{\f}, \MDRG_{\g}\right)$ and $d_{\MDRG,r}^{0}\left(\MDRG_{-\f}, \MDRG_{-\g}\right)$ respectively compute the distance between persistence diagrams of sub-level set and super-level set filtrations of the Reeb graphs in the MDRGs and  $d_{\MDRG,r}^{1}\left(\MDRG_{\f}, \MDRG_{\g}\right)$ is the distance between the $1$-st extended persistence diagrams of the Reeb graphs in $\MDRG_{\f}$ and $\MDRG_{\g}$ respectively. We note, $d_{T,r}$ is a pseudo-metric.
\begin{theorem}
$d_{T,r}$ satisfies the properties of a pseudo-metric.    
\end{theorem}
The proof of this theorem is similar to that of Theorem \ref{thm:psuedo-metric}. Next, we show the stability of $d_{T,r}$.
\begin{theorem}
\textbf{Stability:}
\begin{small}
\begin{equation*}
\begin{aligned}
 d_{T,r}(\MDRG_{\f},\MDRG_{\g}) &\leq 3\|f_1 - g_1\|_{\infty} +\\
 &\indent \frac{1}{2}\sum_{i=2}^{r}\max\{Amp(f_i), Amp(g_i)\}.
\end{aligned}
\end{equation*}
\end{small}
\end{theorem}
The bound on $d_{T,r}(\MDRG_{\f},\MDRG_{\g})$ consists of two terms. The proof of the first and second terms are similar to the proofs of Lemmas \ref{stability-lemma-1} and \ref{stability-lemma-2} respectively.

Next, we show the experimental results of the proposed method on two datasets.

\section{Experimental Results}
\label{sec:experimental-results}
In this section, we show the application of the proposed distance measure in shape matching and detecting topological features in a time varying multi-field data.

\subsection{Classification and Analysis of Shape Data}
\label{subsec:shape-matching}
In this subsection, we show the effectiveness of the proposed distance between MDRGs in matching shapes. First, we describe the feature descriptors corresponding to a shape for computing the MDRGs.

\subsubsection{Feature Descriptors of a Shape}
\label{subsubsec:feature-descriptors-of-shape}
We obtain scalar fields based on the eigenfunctions of discrete Laplace-Beltrami operator (see equation  (\ref{eqn:GeneralizedEigenvalueProblem})) for computing the MDRG.  The eigenvalues $\lambda_i$s of the discrete LB operator  are sorted in the non-decreasing order. The first eigenvalue $\lambda_0$ is zero  and the corresponding  eigenfunctions are constant \cite{2006-Reuter-Shape-DNA}. Therefore, we consider the eigenvalues $\lambda_i$ for $i > 0$. The ability  of the eigenfunctions in capturing the geometric and topological properties of a shape and their invariance to isometry of shapes \cite{2006-Levy-Laplace-Beltrami-Eigenfunctions, 2007-Rustamov-Laplace-Beltrami-Eigenfunctions} make them suitable shape descriptors.

To remove the effect of scale, each eigenfunction is normalized by its corresponding eigenvalue \cite{2007-Rustamov-Laplace-Beltrami-Eigenfunctions}. For a triangulated shape with vertices $v_1,v_2,...,v_n$, the normalized eigenfunction corresponding to eigenvalue $\lambda_i$ is an $n$-dimensional vector $\hat{\phi_i} = \left(\frac{\phi_i(v_1)}{\sqrt{\lambda_i}}, \frac{\phi_i(v_2)}{\sqrt{\lambda_i}}, ..., \frac{\phi_i(v_{n})}{\sqrt{\lambda_i}}\right)$, where $\phi_i(v_j)$ is the value of the $i^{th}$ eigenfunction at the vertex $v_j$.

\begin{figure}
\centering
\includegraphics[width=2.5cm]{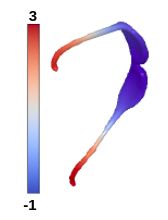}
\qquad
\includegraphics[width=2.4cm]{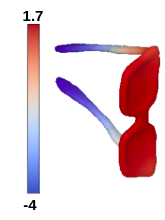}
\caption{Eigenfunction $\phi_3$ of similar shapes}
\label{fig:sign-ambiguity-problem}
\end{figure}

\textbf{Sign Ambiguity Problem: } We now encounter the sign ambiguity problem while dealing with the eigenfunctions of the LB operator. If $\phi_i$ is an eigenfunction corresponding to the  eigenvalue $\lambda_i$ , then $-\phi_i$ is also an eigenfunction for $\lambda_i$. We have to pick one among them as the eigenfunction for $\lambda_i$. If we arbitrarily choose either of $\phi_i$ or $-\phi_i$, then there is a possibility of two similar shapes having eigenfunctions which are negative of each other (See \figref{fig:sign-ambiguity-problem}). Therefore, for each eigenvalue $\lambda_i$, we need to systematically pick one of $\phi_i$ or $-\phi_i$. This is known as the sign ambiguity problem \cite{2007-mateus-shape-matching}, for which Umeyama $\etal$ \cite{1988-Umeyama-EigenDecomposition}  proposed a solution by taking the absolute values of eigenfunctions. In this paper, we adopt this solution and choose the functions $\desc{i}$ as shape descriptors.

We show the effectiveness of using absolute values of eigenfunctions as feature descriptors with respect to HKS \cite{2009-Sun-HKS}, WKS \cite{2011-Aubry-WKS} and SIHKS \cite{2010-Bronstein-SIHKS}. The descriptors HKS and SIHKS depend on the number of  time parameters \cite{2009-Sun-HKS,2010-Bronstein-SIHKS} and WKS depends on the number of energy distributions \cite{2011-Aubry-WKS}. Following \cite{2014-Li-Persistence-Based-Structural-Recognition}, we have fixed the number of time parameters as $10$ for HKS, $17$ for SIHKS, and the number of energy distributions as $10$ for WKS. The number of eigenfunctions is chosen as $200$ for constructing HKS, WKS and SIHKS. The distance between two shapes with respect to a distance measure $d$ is defined as the sum of distances computed at each time (for HKS, SIHKS) or energy distribution (for WKS). For example, the distance between shapes $S_1$ and $S_2$ using HKS is given by:
\begin{small}
\begin{equation}
\label{eqn:Extension-HKS}
    d_{\mathrm{HKS}} = \sum_{t=1}^{10} d(h_t(S_1), h_t(S_2))
\end{equation}
\end{small}
where $d$ is the distance measure and $h_t$ is the HKS at time $t$.

\subsubsection{SHREC $2010$ Dataset}
\label{subsubsec:SHREC-2010}
\begin{figure}
    \centering
    \includegraphics[width=6cm]{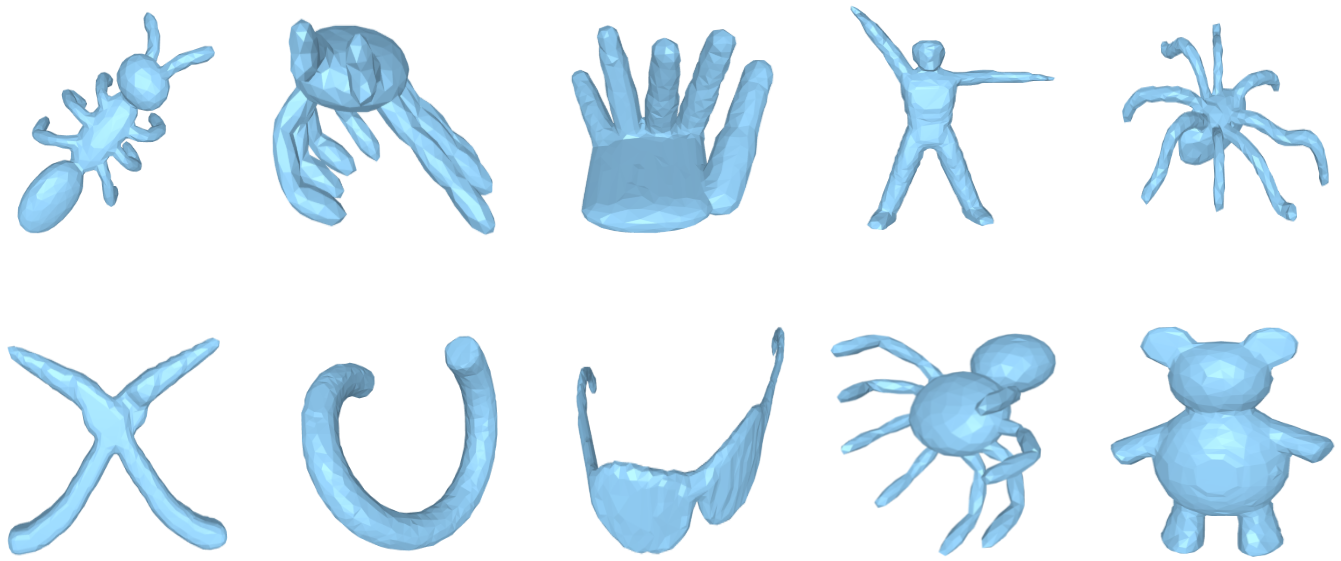}
    \caption{Collection of shapes from SHREC $2010$}
    \label{fig:SHREC-2010-Shapes}
\end{figure}

We show the experimental results of the proposed measure in the Shape Retrieval Contest (SHREC) $2010$ dataset and compare the results with various topological distance measures in the literature. The SHREC $2010$ dataset \cite{2010-SHREC} consists of $200$ watertight shapes classified into $10$ categories. A sample of the shapes are shown in Figure \ref{fig:SHREC-2010-Shapes}. Each of the meshes is simplified into $2000$ faces \cite{2014-Li-Persistence-Based-Structural-Recognition}. We compare the performance of the proposed distance between MDRGs with the distance between MRSs \cite{2021-Ramamurthi-MRS} and the distance between histograms corresponding to fiber-component distributions \cite{2019-Agarwal-histogram}. We evaluate the efficiency of distance measures by the standard evaluation techniques NN, FT, ST, E-measure, and DCG (see \cite{2010-SHREC} for details). These techniques take a distance matrix consisting of distances between all pairs of shapes and return a value between $0$ and $1$, where higher values indicate better efficiency.

We first show the significance of bivariate fields over scalar fields using the proposed distance measure $d_{T}$. Table \ref{table:significance-of-bivaraite-field} shows the results of $d_{T}$ between MDRGs using a pair of eigenfunctions (bivariate field) and the bottleneck distance between Reeb graphs using individual eigenfunctions (scalar field). The bivariate field using MDRG produces a better result than the Reeb graphs of the individual scalar fields, thereby indicating its significance.
\begin{table}[H]
    \centering
    \renewcommand{\arraystretch}{1.5}
    \caption{Performance Results. Comparison of the the proposed distance between MDRGs for the bivariate field $\left(\desc{1},\desc{2}\right)$ with the bottleneck distance between Reeb graphs corresponding to the scalar fields $\desc{1}$ and $\desc{2}$, respectively}
    \begin{scriptsize}
    \begin{tabular}{|c|c|c|c|c|c|}
        \hline
        Descriptor & NN & $1$-Tier & $2$-Tier & e-Measure & DCG\\
        \hline
        $\desc{1}$ & $0.8100$ & $0.5461$ & $0.6950$ & $0.4963$ & $0.8172$\\
        \hline
        $\desc{2}$ & $0.7800$ & $0.4437$ & $0.5924$ & $0.4131$ & $0.7764$\\
        \hline
        $\left(\desc{1},\desc{2}\right)$ & $0.8600$ & $0.5784$ & $0.7187$ & $0.5159$ & $0.8412$\\
        \hline
\end{tabular}
\end{scriptsize}
\renewcommand{\arraystretch}{1}
\label{table:significance-of-bivaraite-field}
\end{table}

The efficiency in computing the distance between shapes will be higher when the shapes are compared based on many eigenfunctions. However, a $3$D shape is a $2$-dimensional manifold. If more than $2$ fields are applied on a shape, then the corresponding Reeb space is not defined. Therefore, we generalize the distance between shapes $S_1$ and $S_2$ for more than two eigenfunctions as follows.
\begin{small}
\begin{equation}
    d(S_1,S_2) = \sum_{i=1}^{E-1} d_{T}\left(\MDRG_{\desc{i}, \desc{i+1}}(S_1), \MDRG_{\desc{i}, \desc{i+1}}(S_2)\right)
\label{eqn:bivariate-field-multiple-eigenfunctions}
\end{equation}
\end{small}
where $E$ is the number of eigenfunctions and $\MDRG_{\desc{i}, \desc{i+1}}(S_k)$ is the MDRG for the shape $S_k$ with $\desc{i}$ and $\desc{i+1}$ as the component scalar fields. Table \ref{table:VaryingNumberOfEigenFunctions-1} shows the performance of the proposed method for varying number of eigenfunctions. The results clearly indicate the increase in efficiency with increase in the number of eigenfunctions.
\begin{small}
\begin{table}[H]
    \centering
    \renewcommand{\arraystretch}{1.5}
    \caption{Performance of the proposed distance between MDRGs for varying number of eigenfunctions}
    \begin{tabular}{|c|c|c|c|c|c|}
        \hline
        $N$ & NN & $1$-Tier & $2$-Tier & e-Measure & DCG\\
        \hline
        $2$ & $0.8600$ & $0.5784$ & $0.7187$ & $0.5159$ & $0.8412$\\
        \hline
        $4$ & $0.9200$ & $0.6318$ & $0.7576$ & $0.5435$ & $0.8700$\\
        \hline
        $6$ & $0.9450$ & $0.6729$ & $0.8058$ & $0.5735$ & $0.8920$\\
        \hline
        $8$ & $0.9450$ & $0.7000$ & $0.8416$ & $0.5963$ & $0.9127$\\
        \hline
        $10$ & $0.9650$ & $0.7518$ & $0.8695$ & $0.6263$ & $0.9367$\\
        \hline
        $12$ & $0.9650$ & $0.7587$ & $0.8734$ & $0.6308$ & $0.9410$\\
        \hline
\end{tabular}
\renewcommand{\arraystretch}{1}
\label{table:VaryingNumberOfEigenFunctions-1}
\end{table}
\end{small}
Table $\ref{table:comparison-of-descriptors}$ shows the performance of HKS, WKS, SIHKS and absolute values of eigenfunctions as shape descriptors for various distance measures. The distance measures for HKS, WKS and SIHKS are computed using equation (\ref{eqn:Extension-HKS}). We set the parameters for various distance measures as follows. The distance between MDRGs is computed as specified in equation (\ref{eqn:bivariate-field-multiple-eigenfunctions}), where $E = 12$ and the weights $w_0, w_1$ and $w_2$ are set equally. The number of slabs for constructing the JCNs (required for computing MDRGs) is $32$ , the parameters $w$ and $q$ for the distance between fiber-component distributions are set to $1$ and the MRSs are constructed for $6$ resolutions.
\begin{table}[H]
    \setlength\tabcolsep{3pt}
    \renewcommand{\arraystretch}{1.5}
    \caption{Performance Results for SHREC $2010$ dataset. Comparison of distance measures based on fiber-component distributions \cite{2019-Agarwal-histogram}, MRSs \cite{2021-Ramamurthi-MRS}, persistence diagrams (PDs) of Reeb graphs \cite{2014-Bauer-DistanceBetweenReebGraphs} and PDs of MDRGs using the descriptors HKS, WKS, SIHKS and pairs of eigenfunctions}
    \begin{tiny}
    \begin{tabular}{|p{1.3cm}|c|c|c|c|c|c|}
    \hline
         Descriptors & Methods & NN & $1$-Tier & $2$-Tier & e-Measure & DCG\\
        \hline
        \multirow{3}{*}{HKS} & Histogram & $0.9200$ & $0.5084$ & $0.6537$ &    $0.4588$ & $0.8200$\\
        \cline{2-7}
        & MRS & $0.9250$ & $0.6821$     & $0.8211$ &    $0.5965$ &       $0.9001$\\
        \cline{2-7}
        & $\PD$ of Reeb graph & $0.9600$ & $0.5805$ & $0.7208$ & $0.5112$ & $0.8695$\\
        \hline
        \multirow{3}{*}{WKS} & Histogram &  $0.9400$ & $0.5008$ & $0.6453$ &    $0.4541$ & $0.8250$\\
        \cline{2-7}
        & MRS & $0.9350$ & $0.5105$     & $0.6468$    & $0.4604$    &    $0.8273$\\
        \cline{2-7}
        & $\PD$ of Reeb graph &  $0.8950$ & $0.4124$ &     $0.5234$    & $0.3731$ &       $0.7665$\\
        \hline
        \multirow{3}{*}{SIHKS} & Histogram & $0.9600$ & $0.7534$ & $0.9008$ & $0.6522$  & $0.9380$\\
        \cline{2-7}
        & MRS & $0.9600$ & $0.7576$   &   $0.8808$ &    $0.6412$  &      $0.9354$\\
        \cline{2-7}
        & $\PD$ of Reeb graph & $0.9750$ & $0.8258$    &  $0.9605$     & $0.6935$ &       $0.9618$\\
        \hline
        \multirow{3}{1cm}{Pairs of Eigen-functions} & Histogram & $0.8900$ & $0.6153$ & $0.7697$ & $0.5469$ & $0.8692$\\
        \cline{2-7}
        & MRS &  $0.9700$ & $0.7774$  &    $0.8821$  &   $0.6431$  & $0.9438$\\
        \cline{2-7}
        & $\PD$s of MDRG & $0.9650$ & $0.7587$ & $0.8734$ & $0.6308$ & $0.9410$\\
        \hline
    \end{tabular}
    \end{tiny}
    \label{table:comparison-of-descriptors}
\end{table}

From the table, it can be seen that MDRG using eigenfunctions produces better results than HKS and WKS (for all the methods). With eigenfunctions as shape descriptors, the proposed method performs better than the histogram of fiber-component distributions and the results are comparable with that of MRS. On the other hand, the bottleneck distance between Reeb graphs outperforms other methods using SIHKS. However, the proposed measure using eigenfunctions performs better than the bottleneck distance between Reeb graphs using SIHKS when we experiment with $7$ categories of shapes, which can be observed in the distance matrices in \figref{fig:7-distance-matrices-7-shapes}.

\subsubsection{Results for Seven Categories of Shapes}
\label{subsubsec:SHREC-2010-7-categories}
\begin{figure}
    \centering
    \includegraphics[width=6cm]{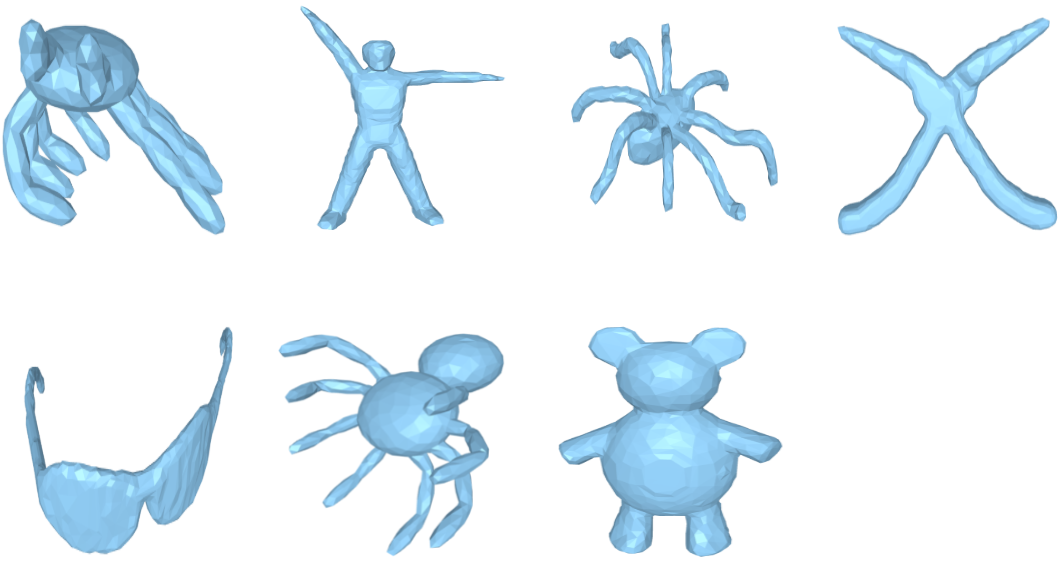}
    \caption{Seven categories of shapes from SHREC $2010$}
    \label{fig:7-shapes-SHREC-2010}
\end{figure}

\begin{figure}
    \centering
    \includegraphics[width=0.47\textwidth]{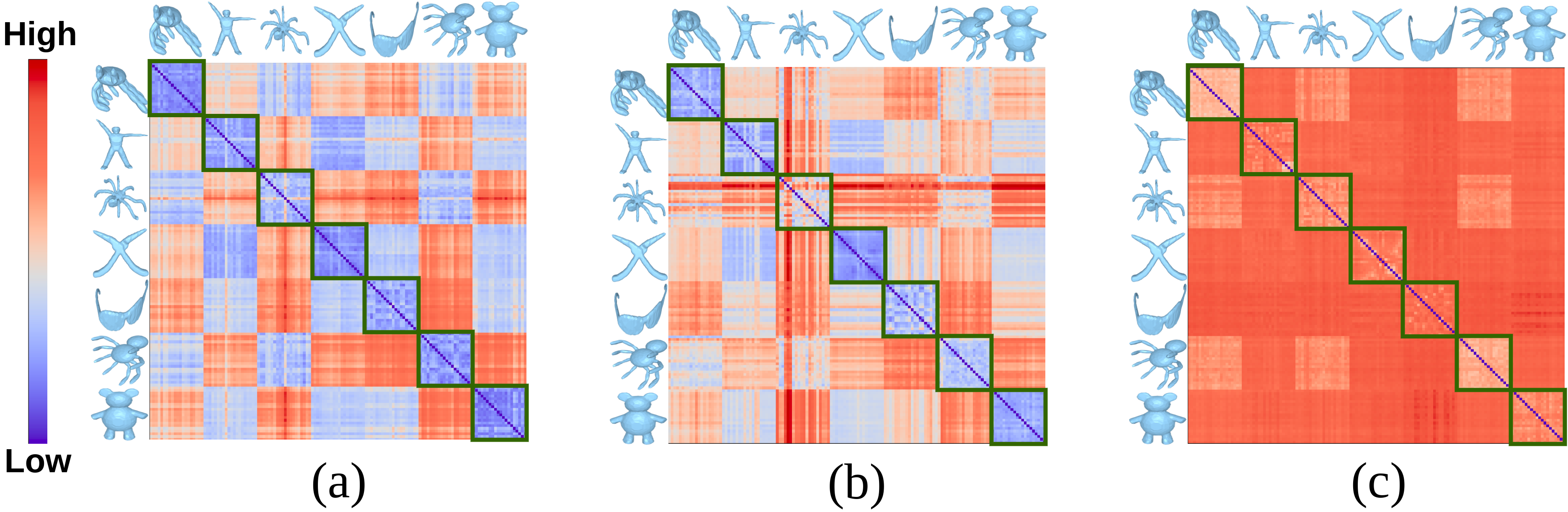}
    \caption{Distance matrices. (a) Bottleneck distance between Reeb graphs using SIHKS. (b), (c) Distances between MDRGs and MRSs, respectively using pairs of eigenfunctions}
    \label{fig:7-distance-matrices-7-shapes}
\end{figure}

We experiment with $7$ of the $10$ categories of shapes shown in \figref{fig:7-shapes-SHREC-2010} and the results are provided in Table \ref{table:7-shapes-comparison-of-descriptors}. Similar to the results in Table \ref{table:comparison-of-descriptors}, MDRG using eigenfunctions performs better than other techniques using HKS and WKS. Further, the results of the distance between MDRGs are comparable with that of MRSs using eigenfunctions and the bottleneck distance between Reeb graphs using SIHKS. This can be seen from the distance matrices shown in \figref{fig:7-distance-matrices-7-shapes}. The proposed distance using MDRGs is better in discriminating different classes of shapes as compared to the distances based on Reeb graphs and MRSs.

\begin{table}
\centering
\renewcommand{\arraystretch}{1.5}
\setlength\tabcolsep{3pt}
    \caption{Performance Results for seven categories of shapes in SHREC $2010$ dataset. Comparison of distance measures based on fiber-component distributions \cite{2019-Agarwal-histogram}, MRSs \cite{2021-Ramamurthi-MRS}, PDs of Reeb graphs \cite{2014-Bauer-DistanceBetweenReebGraphs} and PDs of MDRGs using HKS, WKS, SIHKS and pairs of eigenfunctions as shape descriptors}
    \begin{tiny}
    \begin{tabular}{|p{1.3cm}|c|c|c|c|c|c|}
    \hline
        Descriptors & Methods & NN & $1$-Tier & $2$-Tier & e-Measure & DCG\\
        \hline
        \multirow{3}{*}{HKS} & Histogram&$0.9000$&$0.5534$&$0.7440$&$0.5174$&$0.8407$\\
        \cline{2-7}
        & MRS & $0.9071$ & $0.7658$ & $0.9414$ & $0.6709$ & $0.9357$\\
        \cline{2-7}
        & $\PD$ of Reeb graph & $0.9429$ & $0.6335$ & $0.7805$ & $0.5683$ & $0.8787$\\
        \hline
        \multirow{3}{*}{WKS} & Histogram &  $0.9214$ & $0.5470$     & $0.7252$ & $0.5087$ & $0.8517$\\
        \cline{2-7}
        & MRS & $0.9429$ & $0.5703$ & $0.7425$ & $0.5199$ & $0.8632$\\
        \cline{2-7}
        & $\PD$ of Reeb graph & $0.8857$ & $0.4523$ & $0.6004$ & $0.4171$ & $0.7921$\\
        \hline
        \multirow{3}{*}{SIHKS} & Histogram & $0.9643$ & $0.7868$     & $0.9462$ &  $0.6818$ & $0.9526$\\
        \cline{2-7}
        & MRS & $0.9714$ & $0.7992$ & $0.9421$ & $0.6801$ & $0.9546$\\
        \cline{2-7}
        & $\PD$ of Reeb graph & $0.9786$ & $0.8105$ & $0.9560$ & $0.6849$ & $0.9586$\\
        \hline
        \multirow{3}{1cm}{Pairs of Eigen-functions}  & Histogram  & $0.9214$ & $0.6898$ & $0.8492$ & $0.6059$ & $0.9036$\\
        \cline{2-7}
       & MRS &  $0.9714$ & $0.8477$ & $0.9380$ & $0.6882$ & $0.9623$\\
        \cline{2-7}
        & $\PD$s of MDRG & $0.9571$ &  $0.8534$ & $0.9278$ & $0.6793$ & $0.9633$\\
        \hline        
    \end{tabular}
    \end{tiny}
    \renewcommand{\arraystretch}{1}
    \label{table:7-shapes-comparison-of-descriptors}
\end{table}

\subsubsection{Classification Results}
\label{sec:ShapeMatching-Classification}
We measure the effectiveness of the proposed method in the classification of the shapes by performing a $10$-fold cross-validation using a decision tree classifier. The dataset is randomly divided into $10$ folds. The cross-validation is performed for $10$ iterations. In each iteration, one of the folds is taken as the test set and the rest of the folds constitute the training set. For each shape $S$, we consider a feature vector, which consists of the distances from $S$ to every shape in the dataset. The classifier is trained based on the feature vectors of the shapes in the training set. The category of a shape in the test set is determined by the classifier based on its feature vector.

\begin{figure}
    \centering
    \includegraphics[width=0.5\textwidth]{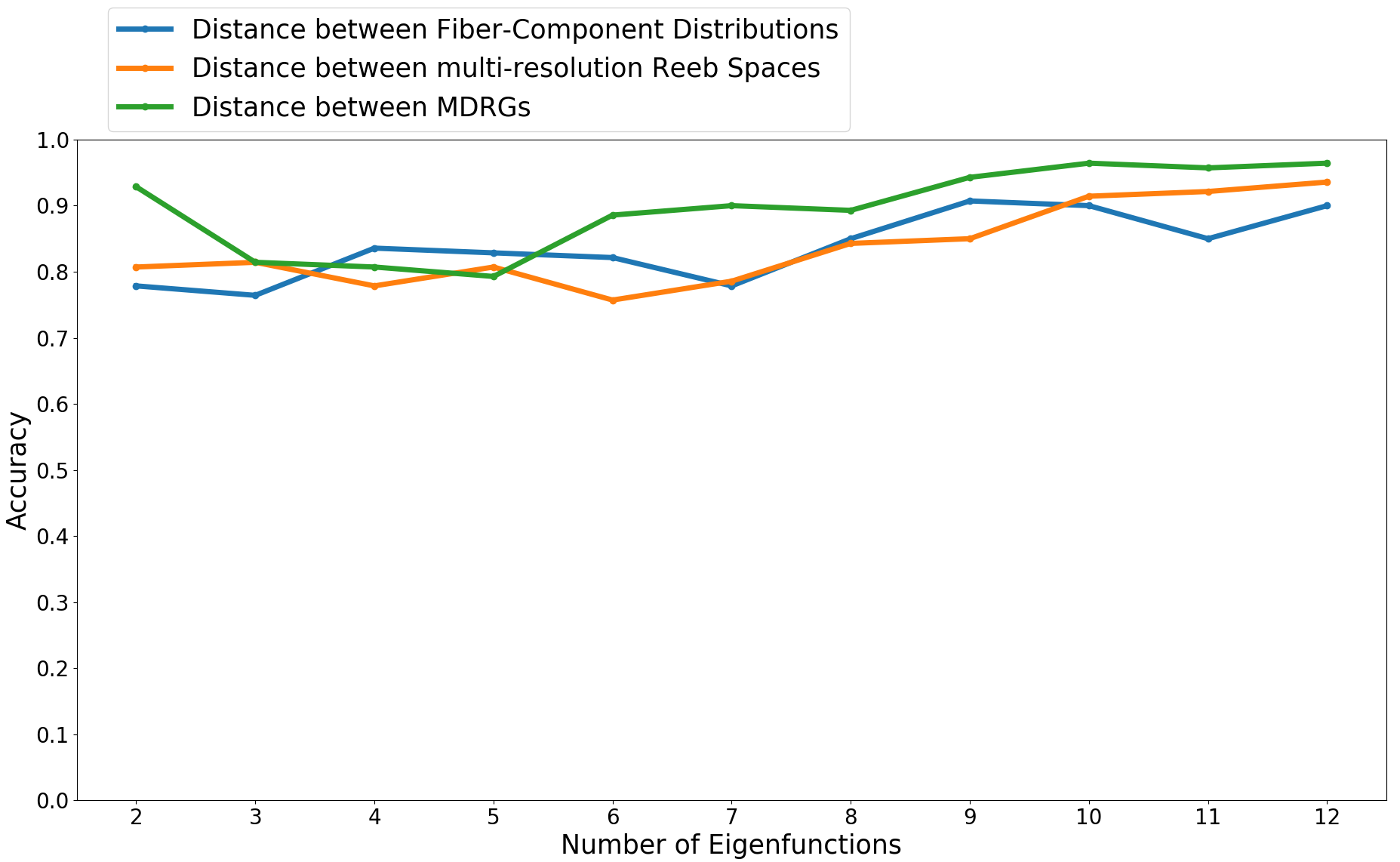}
    \caption{Classification accuracies plotted for varying numbers of eigenfunctions using the multi-field topology-based distances between (i) fiber-component distributions \cite{2019-Agarwal-histogram},  (ii) MRSs \cite{2021-Ramamurthi-MRS}, and (iii) MDRGs. The maximum accuracy ($.96$) is achieved by the proposed distance between MDRGs using $12$ eigenfunctions.}
    \label{fig:classification-accuracy-plots}
\end{figure}

We perform the experiments with $7$ categories of shapes, described in \secref{subsubsec:SHREC-2010-7-categories}. \figref{fig:classification-accuracy-plots} shows the classification accuracy results using three different distances based on multi-field topology, viz., (i) the distance between histograms of fiber-component distributions \cite{2019-Agarwal-histogram}, (ii) the distance between MRSs \cite{2021-Ramamurthi-MRS} and (iii) the proposed distance between MDRGs, using pairs of eigenfunctions as the shape descriptor (as in \secref{subsubsec:SHREC-2010}). The accuracies are plotted for varying numbers of eigenfunctions. From the plots, we observe that the classification accuracy is higher for more eigenfunctions. Further, the proposed distance gives the best accuracy $(96 \%)$ when we use $12$ eigenfunctions as the shape descriptor. 
\begin{figure}
    \centering
    \includegraphics[width=0.5\textwidth]{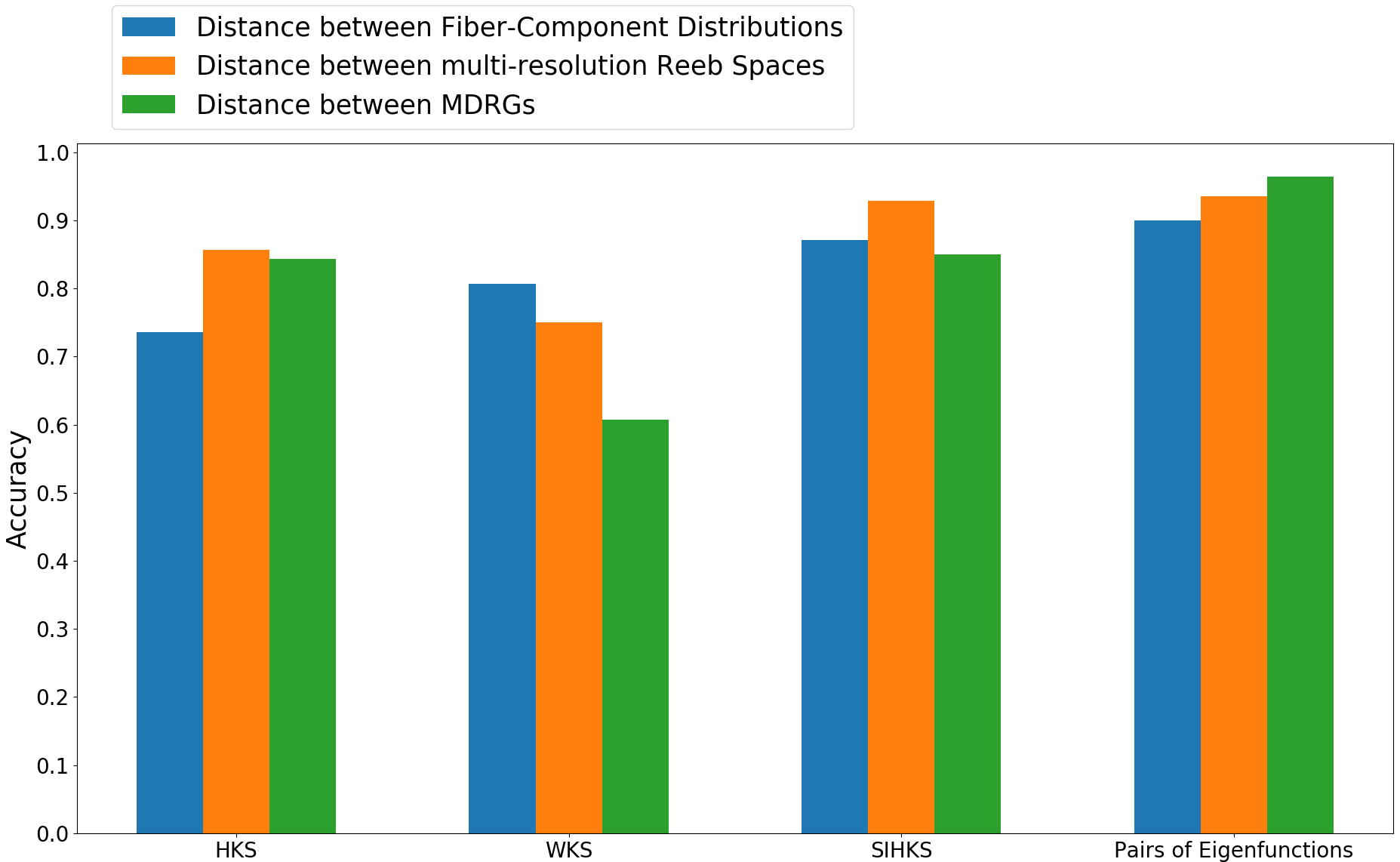}
    \caption{Classification accuracies for the multi-field topology-based distances between (i) fiber-component distributions \cite{2019-Agarwal-histogram},  (ii) MRSs \cite{2021-Ramamurthi-MRS}, and (iii) MDRGs, using the shape descriptors HKS, WKS, SIHKS, and $12$ eigenfunctions (taken in pairs). The MDRG-based method using eigenfunctions achieves the maximum accuracy $(0.96)$.}
    \label{fig:classification-accuracy-plots-descriptors}
\end{figure}

We also evaluate the classification accuracy of the same distances using HKS, WKS, SIHKS, and pairs of eigenfunctions, respectively, as shape descriptors. The results are illustrated in \figref{fig:classification-accuracy-plots-descriptors}. We observe that the eigenfunctions demonstrate a better ability in classifying shapes compared to HKS, WKS, and SIHKS. Further, upon using eigenfunctions as the shape descriptor, the proposed distance between MDRGs shows improved performance over other distances and feature descriptors.

Next, we show the effectiveness of the proposed distance measure in detecting topological features in a time-varying multi-field data from computational chemistry.

\subsection{Classification and Analysis of Chemistry Data}
Adsorption is a process where the molecules of a gas adhere to a metal surface. This phenomenon has various applications including corrosion, electrochemistry, molecular electronics and hetrogeneous catalysis \cite{2010-kendrick-elucidating, 2010-somorjai-introduction}. In particular, the adsorption of the Carbon Monoxide (CO) molecule on the Platinum (Pt) surface has gained interest due to its industrial applications in the areas of automobile emission, fuel cells, and other catalytic processes \cite{2009-dimakis-attraction,2018-patra-surface}. Hence, the interaction between the CO molecule and the platinum surface at the atomic level is studied with utmost importance.

In this study, we validate the proposed method by studying the interaction of the CO molecule with a Pt$_7$ cluster. When the CO molecule approaches the Pt$_7$ cluster, the internal CO bond weakens and a bond is formed between one of the Pt atoms and the C atom of the CO molecule \cite{2009-dimakis-attraction}. The Pt-CO bond is formed at site $13$, which is validated by the geometry in \figref{fig:trivariate-field-Pt-CO}(b). However, the bond length is unstable at this site. The bond stabilizes at site $21$ when the bond length between Pt and C atoms is $1.84$\AA (see the plot in \figref{fig:trivariate-field-Pt-CO}(c)). In this experiment, we aim to capture the stable bond formation between the Pt$_7$ cluster and the CO molecule.

The Pt-CO dataset consists of electron density distributions generated by quantum mechanical computations corresponding to the HOMO (Highest Occupied Molecular Orbital), LUMO (Lowest Unoccupied Molecular Orbital) and HOMO-$1$, defined on a regular $41 \times 41 \times 41$ grid. The orbital numbers $69, 70$ and $71$ correspond to HOMO-$1$, HOMO and LUMO respectively. At a series of $40$ sites, the electron density distributions are computed for varying distances between the C atom of the CO molecule and the Pt surface \cite{2019-Agarwal-histogram}.

\subsubsection{Feature Detection Results}
\begin{figure}
    \centering
    \includegraphics[width=8cm]{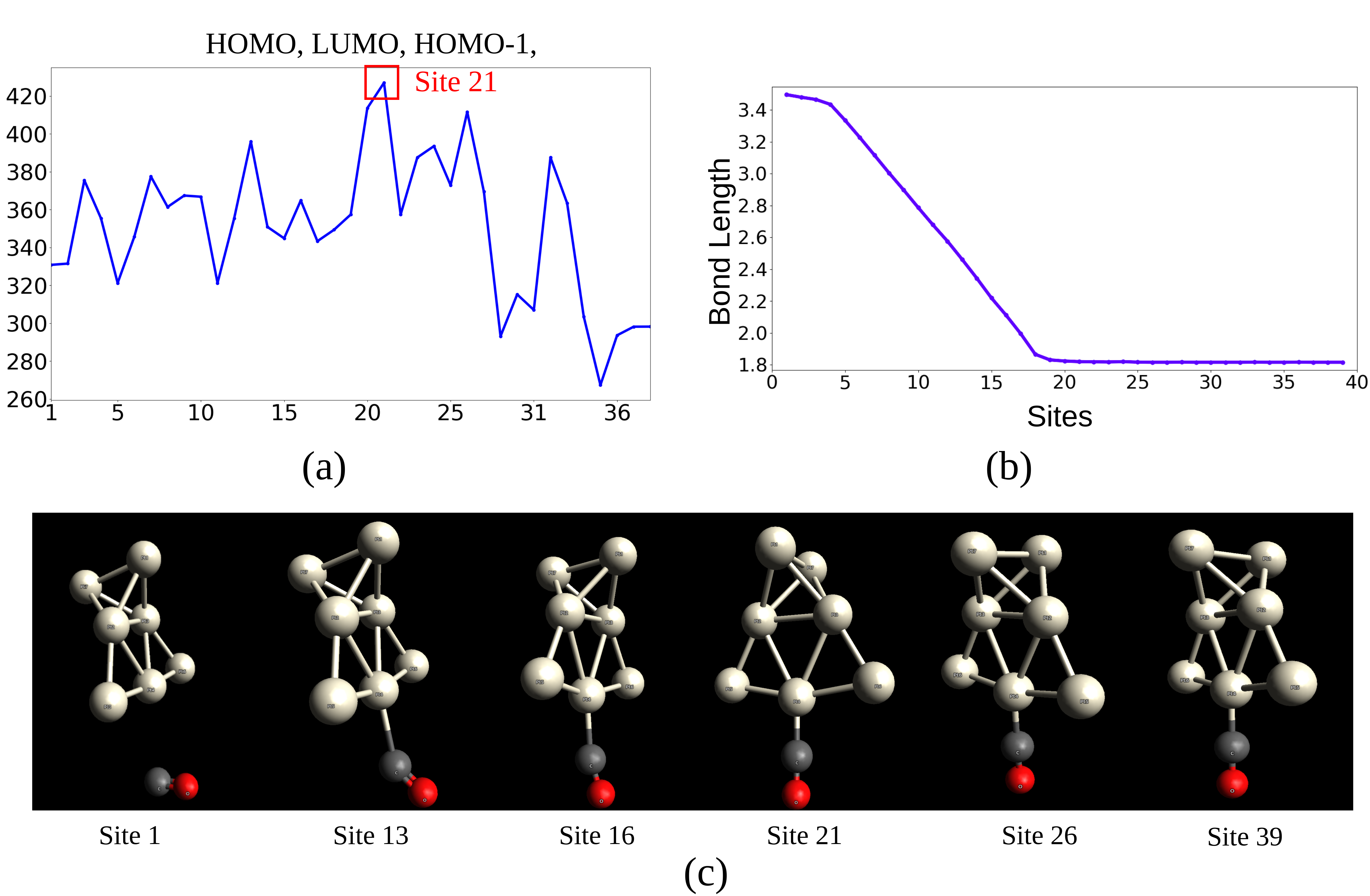}
    \caption{Plots for bond detection in Pt-CO data. (a) Distance between consecutive sites of the data computed using the proposed distance measure for the trivariate field (HOMO, HOMO-$1$, LUMO). The most significant peak is at site $21$, which corresponds to the stable bond formation. (b) Plot of the Pt-CO bond length at various sites. The bond length between Pt and C atoms stabilizes at site $21$. (c) Geometry of the bond formation between the Pt atom and CO molecule. The bond is formed at site $13$ but the bond length is unstable at this site}
    \label{fig:trivariate-field-Pt-CO}
\end{figure}

We examine the performance of the proposed measure in detecting the formation of a stable Pt-CO bond for various combinations of the orbital densities at HOMO, LUMO and HOMO-$1$ molecular orbitals. We compare the performance of scalar, bivariate and trivariate fields. In our experiments, we subdivide the range of each of the fields into $8$ slabs for constructing the JCNs. \figref{fig:trivariate-field-Pt-CO}(a) shows the plots for the trivariate field (HOMO, LUMO, HOMO-$1$), consisting of all three scalar fields. We observe the most significant peak at site $21$. This corresponds to the stable bond formation between Pt and C atoms, which is indicated by the bond length between Pt and C remaining constant after site $21$ (see \figref{fig:trivariate-field-Pt-CO}(b)). Ramamurthi \etal \cite{2021-Ramamurthi-MRS} identified this site of the stable bond formation using the trivariate field.

\begin{figure}
    \centering
    \includegraphics[width=8cm]{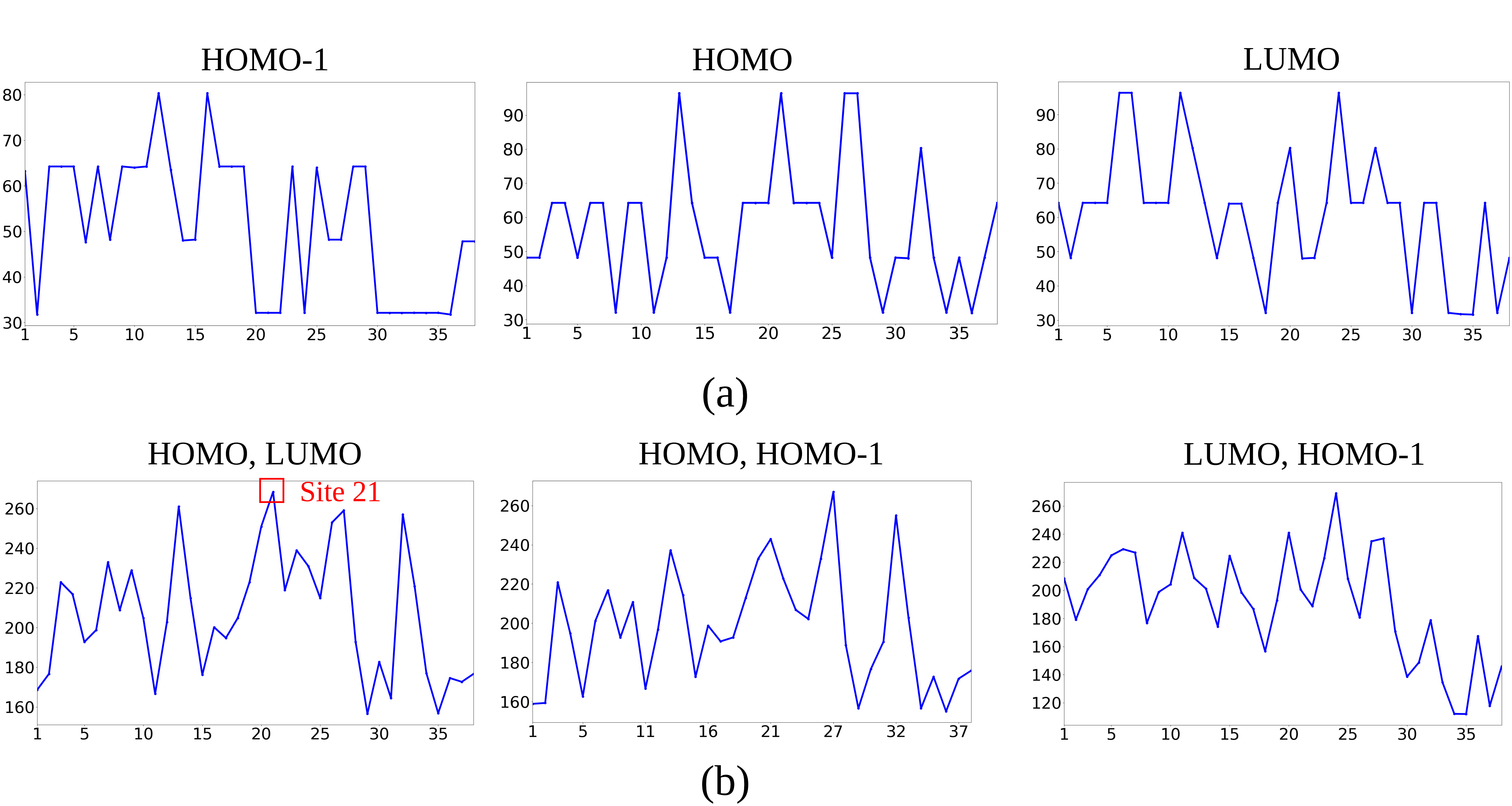}
    \caption{Distance plots for Pt-CO bond detection data using scalar and bivariate fields. (a) Plots for the scalar fields HOMO-$1$, HOMO  and LUMO. (b). Plots for the bivariate fields (HOMO,LUMO), (HOMO, HOMO-$1$) and (LUMO, HOMO-$1$). The stable Pt-CO bond formation is detected using the bivariate field (HOMO, LUMO), which is indicated by the most significant peak at site $21$}
    \label{fig:scalar-bivariate-field-Pt-CO}
\end{figure}

We also measure the performance of the proposed method with scalar fields and bivariate fields. \figref{fig:scalar-bivariate-field-Pt-CO}(a) shows the plots for each of the scalar fields HOMO, HOMO-$1$ and LUMO. We observe that the site of stable bond formation is not detected using each of the scalar fields. \figref{fig:scalar-bivariate-field-Pt-CO}(b) shows the plots for the bivariate fields (HOMO, LUMO), (HOMO, HOMO-$1$) and (LUMO, HOMO-$1$). We observe that the combinations (HOMO, HOMO-$1$) and (LUMO, HOMO-$1$) are unable to detect the correct site in the plots. However, for the combination (HOMO, LUMO), the site corresponding to the stable bond formation is detected, which is also observed in \cite{2019-Agarwal-histogram,2021-Ramamurthi-MRS}. This illustrates the significance of using multi-fields in detecting the bond formation between Pt and C atoms.

\begin{table*}
    \caption{Computational performance results for SHREC $2010$ and Pt-CO datasets. Here, Shapes/Sites: Shapes or sites compared; Field(s): Field(s) for constructing the MDRGs; Slabs: Number of slabs into which the ranges of the component fields are subdivided for constructing the JCNs; $\abs{V_1},\abs{V_2}$: Number of nodes in the JCNs; $t_1$, $t_2$ represent the time (in seconds) for computing the distance between MDRGs based on the $0$th ordinary persistence diagrams corresponding to sub-level set super-level set filtrations respectively and $t_3$ is the time (in seconds) for the computing the distance between MDRGs based on the $1$st extended persistence diagrams}
    \centering
    \begin{tabular}{|c|c|c|c|c|c|c|c|}
        \hline
         Dataset & Shapes/Sites &  Field(s) & Slabs & $\abs{V_1},\abs{V_2}$ & $t_1$ & $t_2$ & $t_3$\\ \hline
         SHREC $2010$  & Human, Hand & $\desc{1}$ & $32, 32$ & $75, 73$ & $0.4556$s & $0.4582$s & $0.4523$s\\ \hline
         SHREC $2010$  & Human, Hand  & $\desc{2}$ & $32, 32$ & $60, 66$ & $0.4040$s & $0.4105$s & $0.3976$s\\ \hline
         SHREC $2010$  & Human, Hand  & $\desc{1},\desc{2}$ & $32, 32$ & $278, 526$ & $0.7789$s & $0.7255$s & $0.7300$s\\ \hline
         Pt-CO & $21, 22$ & HOMO & $8, 8$ & $34, 54$ & $67.1657$s & $67.0570$ & $67.3004$s\\\hline
         Pt-CO & $21, 22$ & HOMO, LUMO & $8, 8$ & $724, 449$ & $79.6311$s & $80.2335$ & $80.7172$s\\\hline
         Pt-CO & $21, 22$ & HOMO, LUMO, HOMO-$1$ & $8, 8$ & $2355, 1649$ & $96.5419$s & $102.9539$ & $95.4901$s\\\hline
    \end{tabular}
\label{table:RuntimeStat}
\end{table*}

\subsubsection{Classification Results}
In this subsection, we evaluate the ability of the proposed distance between MDRGs in classifying a site as occurring before and after the stabilization of the bond between Pt and CO. We divide the dataset into two classes, consisting of the sites before (sites $1$-$21$) and after the stabilization of the Pt-CO bond (sites $22$-$40$), respectively. Similar to \secref{sec:ShapeMatching-Classification}, we employ a decision tree classifier and analyze the effectiveness of methods by performing $10$-fold cross-validation. We measure the classification accuracies and plot the Receiver Operator Characteristics (ROC) curves. The ROC curve is constructed by plotting the true positive rate of classification (TPR) against the false positive rate of classification (FPR). Here, the negative and positive classes are the sites before and after the bond stabilization, respectively. TPR (FPR) gives the proportion of the correct (wrong) predictions in the positive (negative) class. An ROC curve having a higher area under the curve (AUC) indicates better distinguishing ability between the classes.
\begin{figure*}
\centering
\begin{subfigure}{.5\textwidth}
  \centering
  \includegraphics[width=0.9\textwidth]{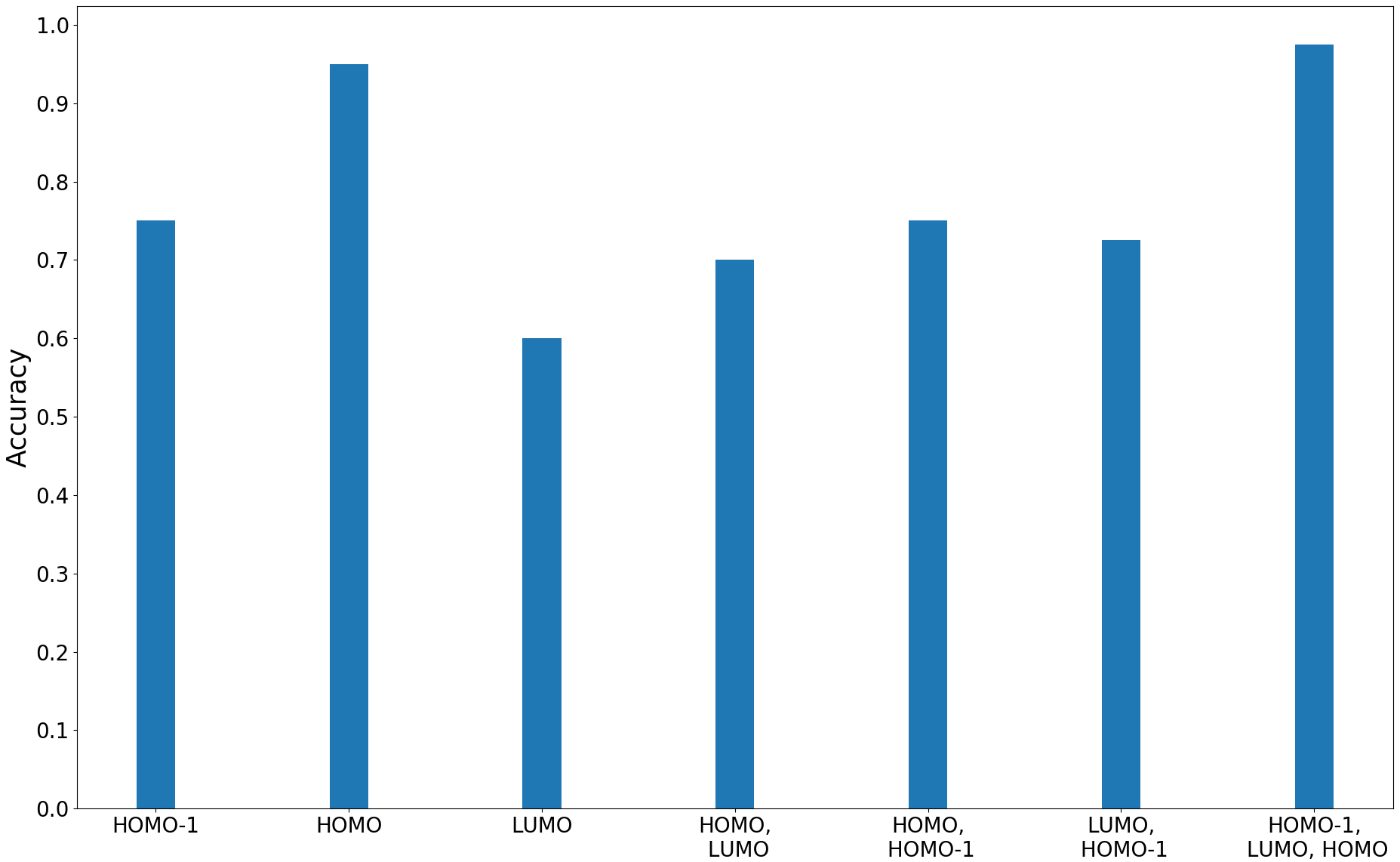}
  \caption{Classification Accuracies}
\end{subfigure}%
\begin{subfigure}{.5\textwidth}
  \centering
  \includegraphics[width=.9\textwidth]{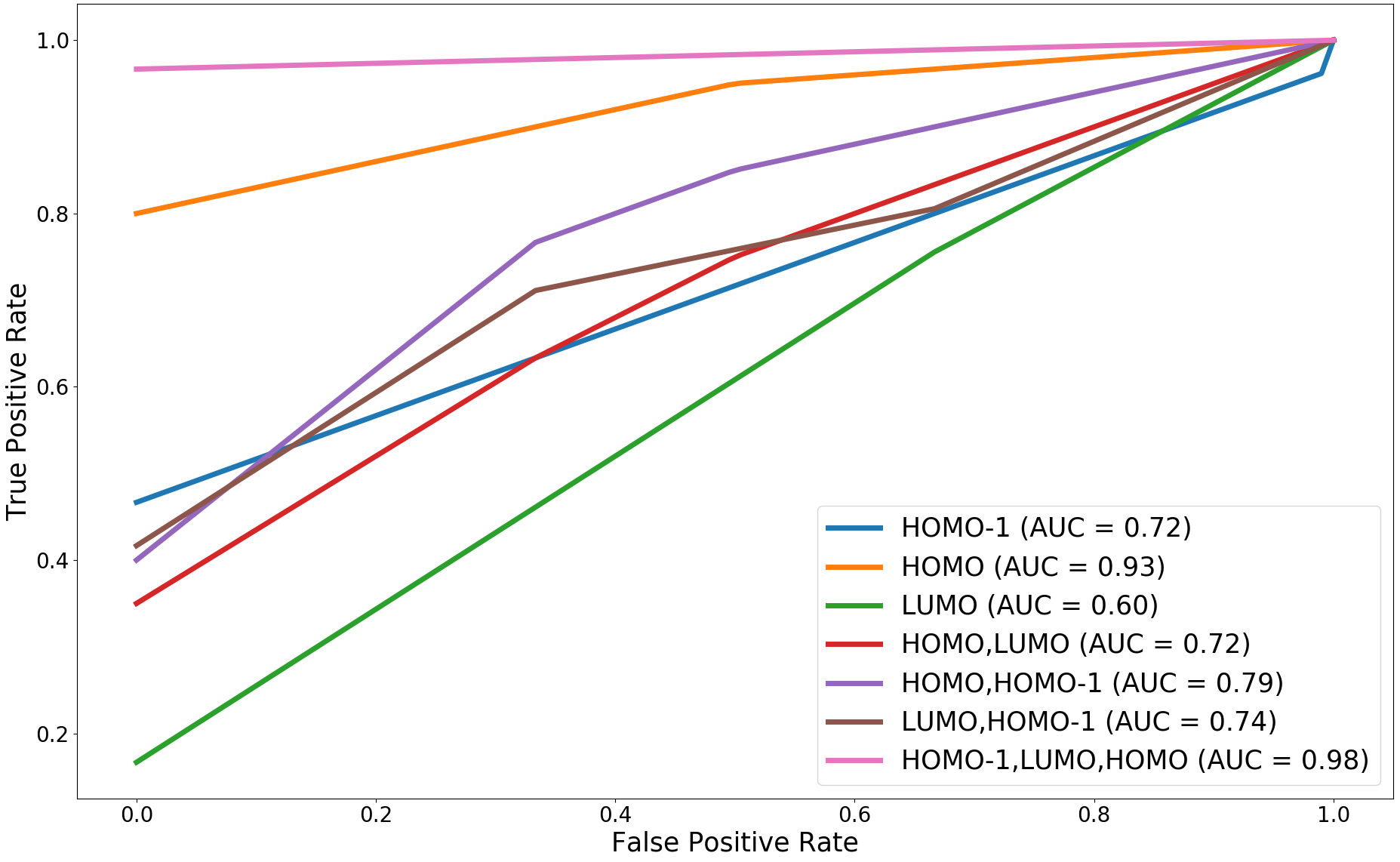}
  \caption{ROC Curve}
\end{subfigure}
\caption{Classification accuracies and ROC curve plots for the proposed distance using the field combinations HOMO, LUMO, LUMO, (HOMO, LUMO), (HOMO, HOMO-$1$), (LUMO, HOMO-$1$) and (HOMO-$1$, LUMO, HOMO). The trivariate field (HOMO-$1$, LUMO, HOMO) has maximum classification accuracy $(0.975)$ and area under the curve $(0.98)$, which indicates better classification ability upon using all the fields.}
\label{fig:DFT-classification-scalar-multi-fields}
\end{figure*}

\figref{fig:DFT-classification-scalar-multi-fields} shows the results of classification accuracies and plots of the ROC curves for the proposed method using various combinations of HOMO, HOMO-$1$ and LUMO fields. Here, the ROC curve is computed as the mean of the ROC curves obtained for each of the $10$ folds used as a test set, in the $10$-fold cross-validation. We observe that the distance between MDRGs using the trivariate field (HOMO-$1$, LUMO, HOMO) obtains a better accuracy and has a higher area under the curve (AUC) compared to other field combinations.

\begin{figure*}
\centering
\begin{subfigure}{.5\textwidth}
  \centering
  \includegraphics[width=0.9\textwidth]{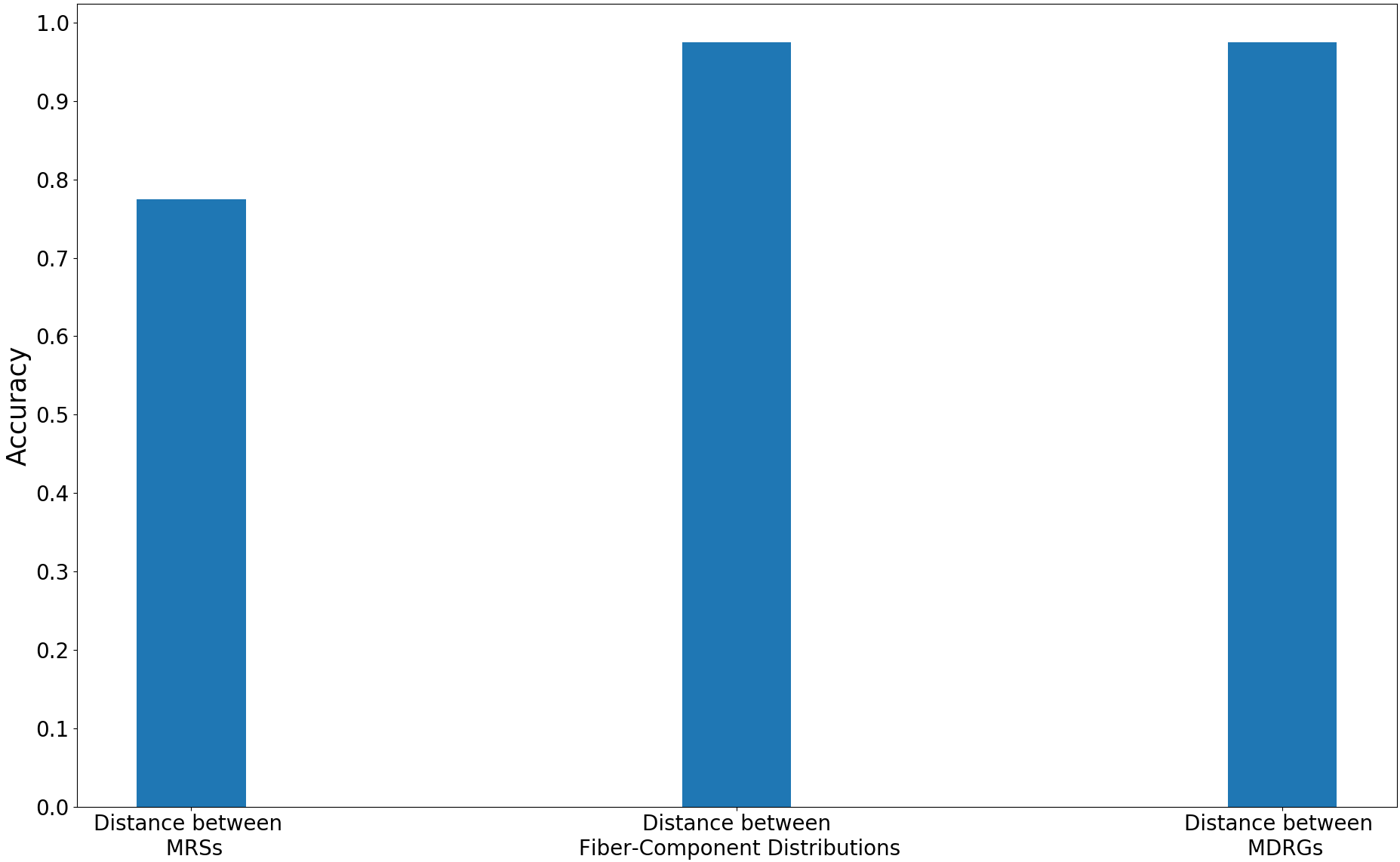}
  \caption{Classification Accuracies}
\end{subfigure}%
\begin{subfigure}{.5\textwidth}
  \centering
  \includegraphics[width=.9\textwidth]{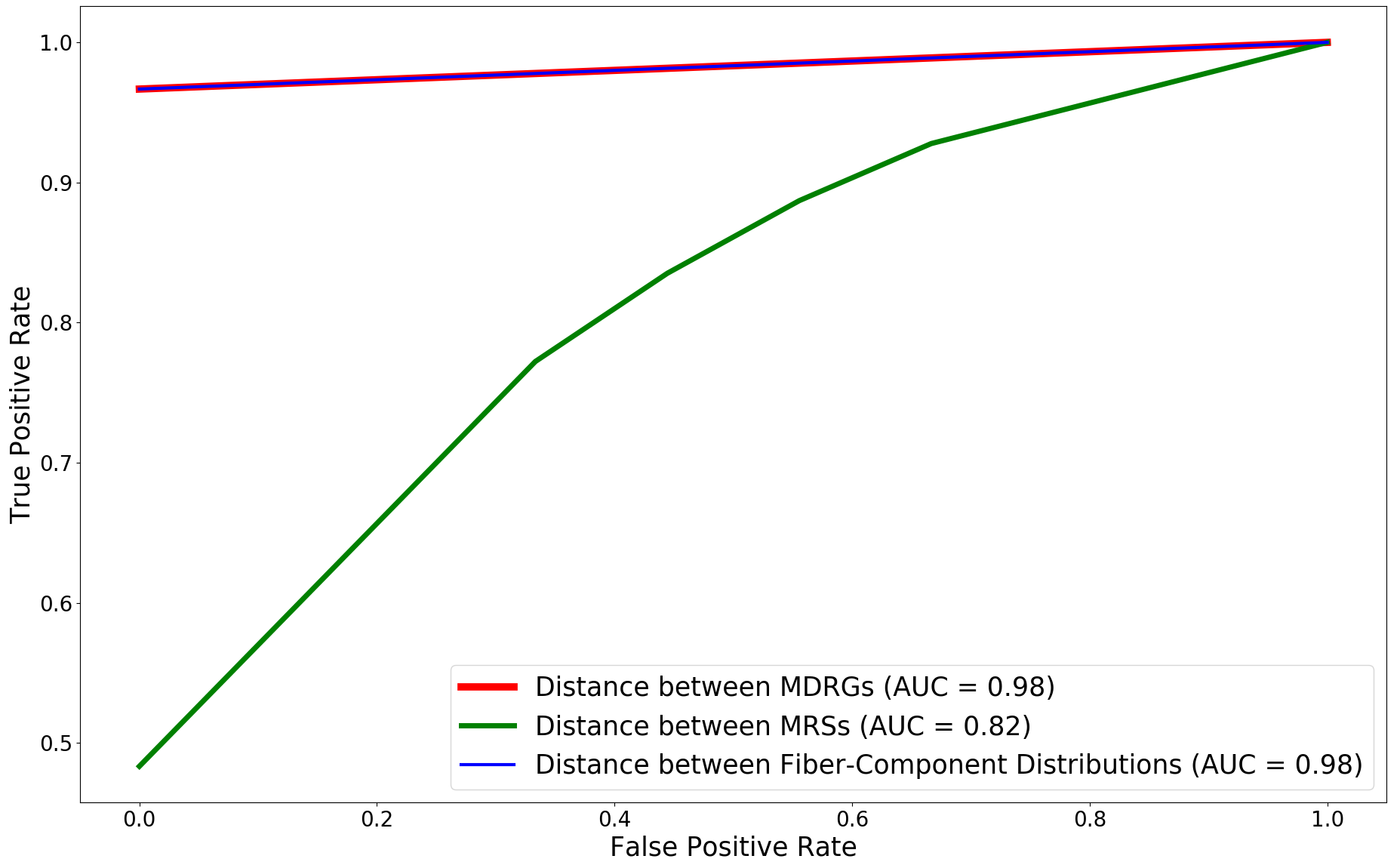}
  \caption{ROC Curve}
\end{subfigure}
\caption{Classification accuracies and ROC curve plots for the distance measures based on fiber-component distributions \cite{2019-Agarwal-histogram}, MRSs \cite{2021-Ramamurthi-MRS}, and MDRGs. The results are computed for the trivariate field (HOMO-$1$, LUMO, HOMO). The MDRG-based distance performs better than the MRS-based distance and its classification ability is same as the fiber-component distribution-based distance.}
\label{fig:DFT-classification-comparison-of-methods}
\end{figure*}
We compare the performance of the proposed method with other multi-field based distances by plotting the classification accuracies and ROC curves for the trivariate field (HOMO-$1$, LUMO, HOMO). The plots are shown in \figref{fig:DFT-classification-comparison-of-methods}. The proposed method performs better than the distance between MRSs \cite{2021-Ramamurthi-MRS} and its performance is on par with the distance between fiber-component distributions \cite{2019-Agarwal-histogram}.

\subsection{Computational Performance}
Table \ref{table:RuntimeStat} shows the computational performance results of the proposed distance between MDRGs in shape matching (SHREC $2010$) and Pt-CO bond detection datasets using scalar and multi-fields. All timings were performed on a $2.20$ GHz $10$-Core Intel Xeon(R) with $16$ GB memory, running Ubuntu $16.04$. From the table, we observe the time taken for computing the distance between MDRGs based on the $0$th ordinary and $1$st extended persistence diagrams respectively. 

\section{Conclusion}
\label{sec:conclusion}
In this article, we introduce a novel distance measure between MDRGs by computing the bottleneck distances between their component Reeb graphs. We show that the proposed distance measure is a pseudo-metric and satisfies the stability property. The effectiveness of the proposed distance measure is shown in the problem of shape classification, where the performance of using absolute values of eigenfunctions as shape descriptors is compared with HKS, WKS and SIHKS. Further, we have shown the performance of the proposed distance measure in detecting the formation of a stable bond between Pt and CO molecules. The ability of the distance is demonstrated in the classification of sites based on their occurrence before and after the stabilization of the Pt-CO bond. 

The computation of the MDRG of a multi-field requires a JCN, which is computationally expensive to construct. Thus, there is a need for a faster algorithm for constructing the MDRG. We have shown applications of the proposed method in the problem of shape retrieval and in detecting the formation of a stable Pt-CO bond, which is an application in the field of computational chemistry. However, exploring other computational domains will be useful for further analysis on the effectiveness of the proposed distance measure. Further, we note, the multiplicity of the eigenvalues of the Laplace Beltrami operator can be greater than $1$, i.e. there can be more than one eigenfunction corresponding to an eigenvalue. In such cases, there is an ambiguity in ordering the eigenfunctions according to eigenvalues. Such issues need to be addressed in the future. 

\section{Declarations}
\textbf{Funding:} The authors would like to thank the Science and Engineering Research Board (SERB), India (SERB/CRG/2018/000702) and International Institute of Information Technology (IIITB), Bangalore for funding this project and for generous travel support.\\
\textbf{Conflicts of interest:} 
The authors of this paper have no conflicts of interest.

\noindent \textbf{Data availability statement:} The datasets analysed during the current study are available from the corresponding author on reasonable request.


\bibliographystyle{spmpsci}

\end{document}